\documentclass[letterpaper, 10 pt, journal, twoside]{IEEEtran}  

\usepackage{color}
\usepackage{graphicx}
\usepackage{amsmath}
\usepackage{amssymb}
\usepackage{tabularx}
\usepackage{subcaption}
\usepackage{url}
\usepackage{cite}

\DeclareMathOperator*{\argmin}{\arg\!\min}

\title{Deep-Learned Collision Avoidance Policy for Distributed Multi-Agent Navigation} 

\author{Pinxin Long, Wenxi Liu*, and Jia Pan*%
\thanks{Manuscript received: September, 10, 2016; Revised December, 6, 2016; Accepted December, 29, 2016. This paper was recommended for publication by Editor Nancy Amato upon evaluation of the Associate Editor and Reviewers' comments.
This work was supported by HKSAR
Research  Grants  Council  (RGC)  General  Research  Fund
(GRF), CityU 17204115, 21203216, and NSFC/RGC Joint Research Scheme CityU103/16 . Asterisk
indicates the corresponding author.}
\thanks{Pinxin Long is with Dorabot Inc., Shenzhen, China. This work was done while the author was at the City University of Hong Kong (e-mail: pinxinlong@gmail.com)}%
\thanks{*Wenxi Liu is with the Department of Computer Science, Fuzhou University, Fuzhou, China (e-mail: wenxi.liu@hotmail.com)}%
\thanks{*Jia Pan is with the Department of Mechanical and Biomedical Engineering, City University of Hong Kong, Hong Kong, China (e-mail: jiapan@cityu.edu.hk)}%
\thanks{Digital Object Identifier (DOI): see top of this page.}
}

\begin{document}
\maketitle

\begin{abstract}
High-speed, low-latency obstacle avoidance that is insensitive to sensor noise is essential for enabling multiple decentralized robots to function reliably in cluttered and dynamic environments. While other distributed multi-agent collision avoidance systems exist, these systems require online geometric optimization where tedious parameter tuning and perfect sensing are necessary. 

We present a novel end-to-end framework to generate reactive collision avoidance policy for efficient distributed multi-agent navigation. Our method formulates an agent's navigation strategy as a deep neural network mapping from the observed noisy sensor measurements to the agent's steering commands in terms of movement velocity. We train the network on a large number of frames of collision avoidance data collected by repeatedly running a multi-agent simulator with different parameter settings. We validate the learned deep neural network policy in a set of simulated and real scenarios with noisy measurements and demonstrate that our method is able to generate a robust navigation strategy that is insensitive to imperfect sensing and works reliably in all situations. We also show that our method can be well generalized to scenarios that do not appear in our training data, including scenes with static obstacles and agents with different sizes. Videos are available at \url{https://sites.google.com/view/deepmaca}.

\end{abstract}


\section{Introduction}

\IEEEPARstart{S}{afe} collision avoidance within multi-agent systems is a fundamental problem in robotics, and has many applications including swarm robotics, crowd simulation, AI games, autonomous warehouse and logistics. The problem can generally be defined in the context of an autonomous agent navigating in a scenario with static obstacles and other moving agents. Each agent needs to compute an action at real time and ensure that by executing the action the agent will not collide with the obstacles and other moving agents while making progress towards its goal.

Previous work about multi-agent navigation can be classified into two categories: centralized and decentralized approaches. The centralized approaches focused on computing time-optimal trajectories for all agents to reach their individual goals in a scene with only static obstacles. These methods solve a large optimization problem to compute the time-optimal plans for all agents simultaneously. For this purpose, they usually have a complete knowledge about all agents' initial and goal states, and require a perfect communication (i.e., with small error and delay) between the agents and a central coordinate controller, which are difficult to achieve in practice. In addition, the centralized planning system is difficult to scale to handle large numbers of agents and are not robust to motion errors as well as agent failures.

\begin{figure}[t] 
\centering
\begin{subfigure}{0.20\textwidth}
\includegraphics[width=1.0\linewidth, height=3.5cm]{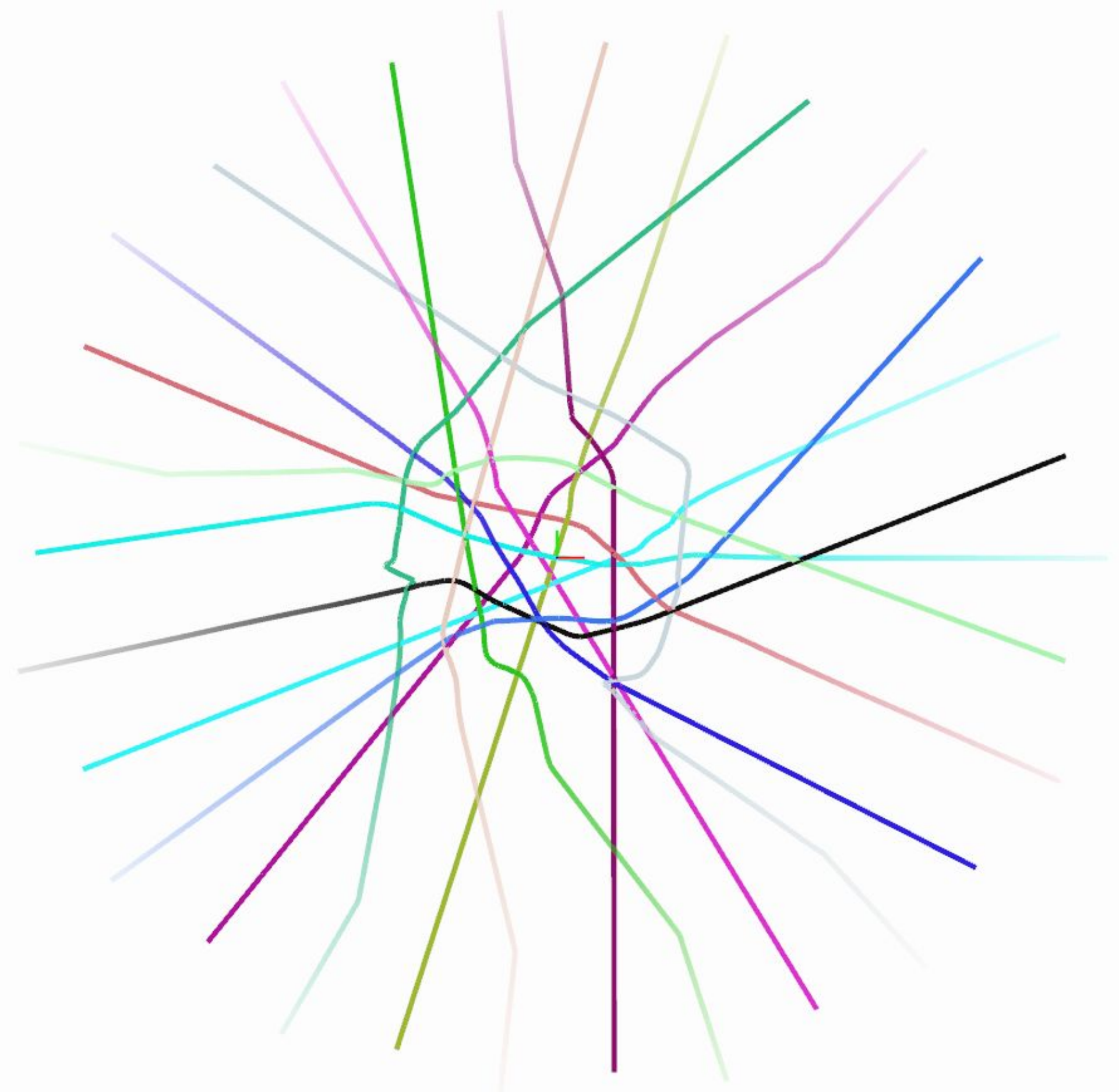}
\caption{ORCA}
\label{fig:circle-rvo}
\end{subfigure}
\begin{subfigure}{0.20\textwidth}
\includegraphics[width=1.0\linewidth, height=3.5cm]{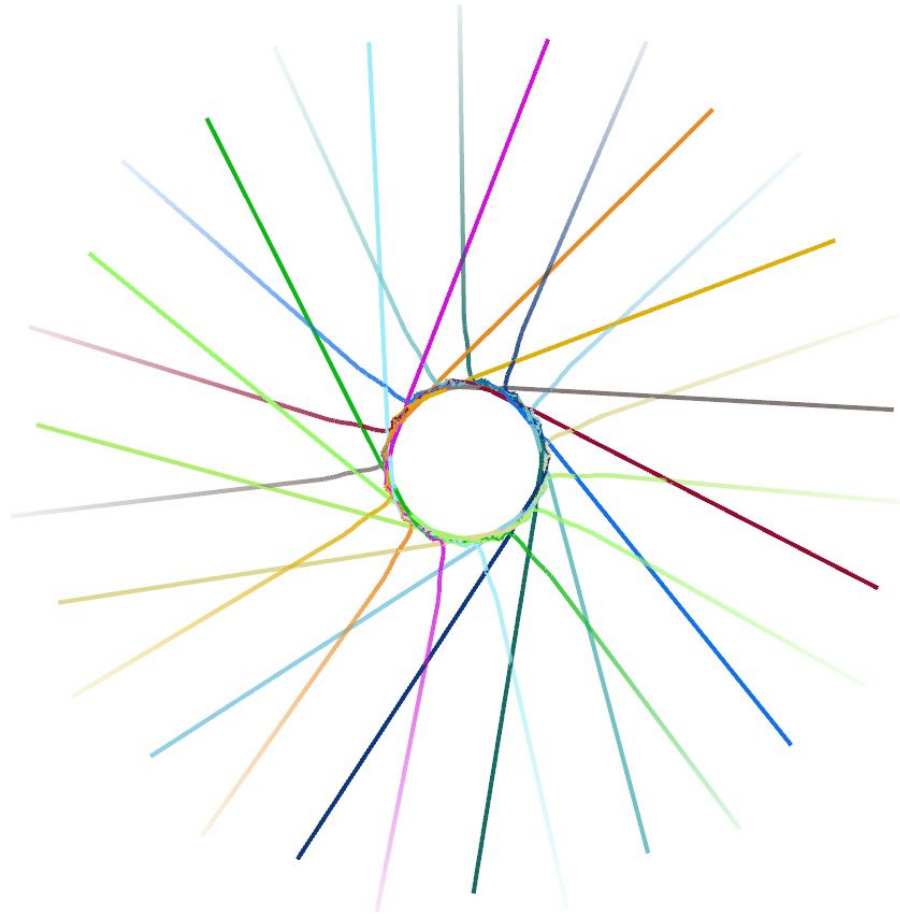}
\caption{Our method}
\label{fig:circle-dnn}
\end{subfigure}

\caption{Agent trajectories in the \emph{Circle} scenario using (a) ORCA method and (b) our learned policy. }
\label{fig:circle}
\vspace*{-0.2in}
\end{figure}

To solve the multi-agent navigation in a decentralized manner, we need to replan each agent's local path at real time to deal with the possible conflict with other agents. Among extensive work addressing this problem, the velocity-based approaches~\cite{van2008reciprocal, Berg:ORCA:2011, snape2011hybrid, hennes2012multi, bareiss2015generalized} have gained in popularity due to their robustness and ability to guarantee local collision-free motion for many agents in a cluttered workspace. In the velocity-based framework, each agent employs a continuous cycle of sensing and acting where the action must be computed according to local observations of the environment. These approaches have two main limitations: First, they assume that each agent has perfect sensing about the surrounding environment, while this assumption may be violated due to sensing uncertainty ubiquitous in the real world. This limitation is alleviated in some previous work by using a global localization system (e.g., an overhead motion capture system) to monitor the positions of all agents~\cite{snape2011hybrid,bareiss2015generalized}, or using an inter-agent communication protocol for sharing position and velocity information among nearby agents~\cite{hennes2012multi,godoy2016implicit,claes2012collision}. Second, velocity-based methods usually have many parameters that are sensitive to the scenario settings (e.g., number of agents and shape of obstacles) and thus must be carefully tuned for achieving satisfactory navigation performance. Unfortunately, there is no systematic principle about how to select these parameters, and the manual parameter tuning is tedious.

These limitations motivate us to develop a novel decentralized collision avoidance technique for multi-agent navigation, which should not only work in real-world settings without perfect sensing and inter-agent communications but should also provide good collision avoidance performance without tedious parameter tuning. 

\noindent \textbf{Main results:} 
We present a learning-based collision avoidance framework that provides an end-to-end solution for distributed multi-agent navigation by directly mapping noisy sensor measurements to a steering velocity that is locally collision-free. The end-to-end framework is implemented as Deep Neural Networks (DNNs).
To train the network, we collect an extensive dataset consisting of frames showing how an agent should avoid its surrounding agents. Each frame includes both the agent's observation about other agents and the agent's reactive collision avoidance strategy in terms of the steering velocity. The dataset is generated by using a state-of-the-art multi-agent simulator with various parameter settings. We also perform data augmentation on the collected dataset by adding measurement noises and leveraging symmetries to generate more frames, which help to reduce over-fitting and improve the generalization capability of our framework.  We train the neural network in an offline manner, and the network learns how to output a collision-avoidance velocity given inputs determined by the agent's sensor measurements and its goal setting. During the online test, our network will output a local collision avoidance velocity that is then used to update the agent's position at each sensing-acting cycle until the agent reaches its goal. We evaluate our approach on a variety of simulation scenarios and compare it to the state-of-the-art distributed collision avoidance framework~\cite{Berg:ORCA:2011}. 
Our experiments show that our method can effectively generate collision-free motions for multiple agents, and the learned collision avoidance policy is robust against noises in the sensor measurements. 
Moreover, we also highlight that the learned policy can be well generalized to scenarios that are unseen in the training data, including scenes with static obstacles and with a different number of agents.


\begin{figure}
\centering
\includegraphics[width=0.6\linewidth]{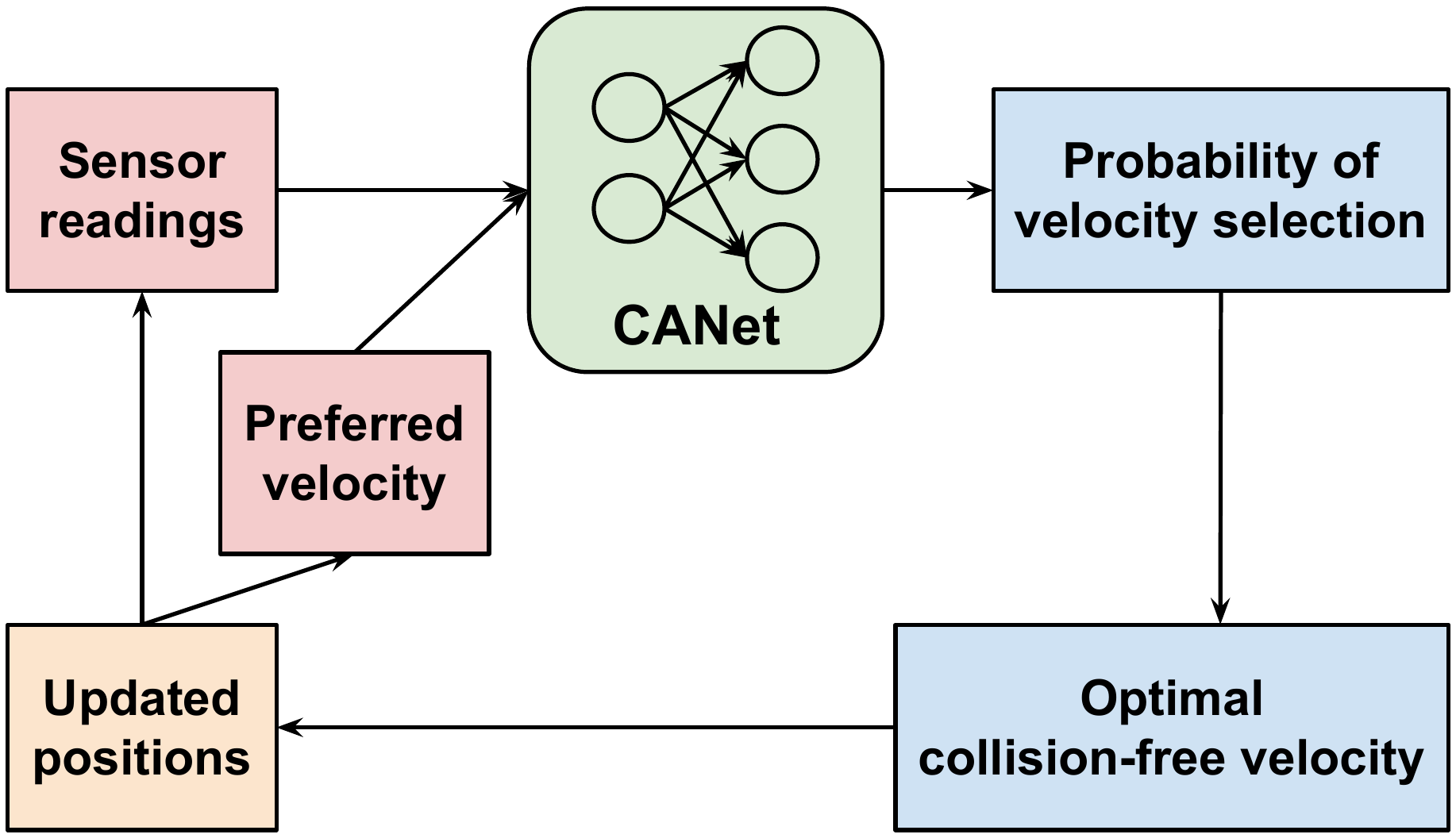} 
\caption{An overview of our approach. A large number of collision avoidance frames collected by repeatedly running a multi-agent simulator are used to train the reactive navigation controller in terms of a collision avoidance network (CANet). At each frame, we record an agent's sensor measurement $\mathbf z$ about the other agents, the agent's preferred velocity $\mathbf v^{pref}$ related to the agent's individual goal, and the corresponding collision avoidance velocity $\mathbf v$; we also estimate $\dot{\mathbf z}$, the velocity about the sensor measurement. All these quantities are then converted into the agent's local coordinate. The navigation controller then has $\mathbf z$, $\dot{\mathbf z}$, and $\mathbf v^{pref}$ as inputs, and $\mathbf v^+$ as the output. During the online navigation, the learned navigation policy is used by agents to
make reactive collision avoidance decisions.}
\label{fig:overview}
\vspace*{-0.2in}
\end{figure}

\section{Related Work}
\label{sec:related}
In this section, we provide a brief overview of prior work on collision avoidance for multi-agent navigation and machine learning for multi-agent systems.  

\subsection{Collision Avoidance for Multi-Agent Navigation}
Collision avoidance has been studied extensively for safe and efficient multi-agent navigation. Many approaches have been proposed, including techniques based on potential fields~\cite{koren1991potential}, local variable-resolution fields~\cite{kapadia2012parallelized}, dynamic windows~\cite{Foxdw}, and velocity obstacles~\cite{fiorini1998motion,van2008reciprocal,Berg:ORCA:2011}.
Among them is the Optimal Reciprocal Collision Avoidance (ORCA) navigation framework~\cite{Berg:ORCA:2011}, which has been a successful velocity-based approach to avoid collisions with other moving agents and obstacles. ORCA has been popular in crowd simulation and multi-agent systems due to its two properties. First, it provides a sufficient condition for multiple robots to avoid collisions with each other, and thus can guarantee collision-free navigation; second, it is a fully distributed method where robots share no knowledge with each other, and thus can easily be scaled to handle large systems with many robots. However, ORCA and its variants (e.g.,~\cite{snape2011hybrid,bareiss2015generalized}) have many parameters that are difficult to tune. More important, these methods are sensitive to the uncertainties ubiquitous in the real-world scenarios. In particular, each robot is assumed to have an accurate observation about the surrounding agents' positions, velocities and shapes; while in practice such information is extracted from noisy sensing measurement via segmentation and tracking, and thus may have significant uncertainties. To alleviate the requirement of perfect sensing, Hennes et al.~\cite{hennes2012multi} and Claes et al.~\cite{claes2012collision} extended the ORCA paradigm with an inter-agent communication protocol for sharing knowledge about agents' positions and
velocities. A one-way communication scheme is also introduced
in~\cite{godoy2016implicit} to coordinate the movement of agents in a crowd scenario. Other approaches~\cite{snape2011hybrid,bareiss2015generalized} avoided
the difficulty in sensing uncertainty by using an overhead motion capture system to obtain the global position observation for all agents. Moreover, Yoon et al.~\cite{yoon2016filling} used multiple visual sensors to track individuals' trajectories in crowds. In this paper, we use
an end-to-end framework to learn a collision-avoidance policy which is robust to the imperfect sensing and requires no inter-agent communication, and thus is still fully distributed. 

\subsection{Machine Learning for Multi-Agent Systems} 
Reinforcement learning has been widely used for the multi-agent decision making~\cite{mataric1997reinforcement,stone2000multiagent,yang2004multiagent,panait2005cooperative}, which is formulated as a multi-agent Markov Decision Processes (MDP) problem. Multi-agent reinforcement learning allows agents to learn a policy, i.e., a mapping from agent states to actions according to the rewards obtained while interacting with the surrounding environment. In these methods, each independent agent needs to build an MDP model via repeated offline simulation, and then uses the model-based reinforcement learning to compute an optimal policy. An online multi-agent policy adaption approach is proposed by Godoy et al.~\cite{godoy2015adaptive}, which originates from multi-arm bandits. They use an online learning framework to plan over the space of preferred velocities and then project these velocities to collision-free ones using ORCA, e.g., their method still holds the perfect sensing assumption and requires parameter-tuning for ORCA. The cooperative approach proposed by Kretzschmar et al.~\cite{kretzschmar2016socially} first infers the internal goals of other agents, then plans a set of jointly possible paths for all neighboring agents in the shared environment. However, it is computationally expensive for generating paths for all others agents. Boatright et al.~\cite{boatright2015generating} train a set of machine-learned policies by decomposing possible scenarios an agent may encounter into steering contexts.
In this paper, we use an end-to-end learning mechanism for supervised training a deep neural network policy from an extensive collection of multi-agent collision avoidance data.

\section{Problem Formulation}
\label{sec:prob}
The multi-agent navigation problem can be formally defined as follows. We take as given a set of $n$ decision-making agents sharing a 2D environment consisting of obstacles. For simplicity, we assume the geometric shape of each agent $a_i$ $(1 \leq i \leq n)$ is modeled as a disc with a fixed radius $r_{a_i}$. In addition, the agent's dynamics is assumed to be holonomic, i.e., it can move in any direction in the 2D workspace. 

Each agent employs a continual cycle of sensing and acting with a time period $\tau$. During each cycle, the agent $a_i$
computes a local trajectory that starts from its current position $\mathbf p_{a_i}$ and has a smallest deviation from a preferred velocity $\mathbf v_{a_i}^{pref}$. The preferred velocity is used to guide the agent in making progress toward its goal $\mathbf g_{a_i}$. In a scenario without static obstacles, $\mathbf v_{a_i}^{pref}$ can directly point toward  $\mathbf g_{a_i}$; in a scene with static obstacles, $\mathbf v_{a_i}^{pref}$ may point toward a closest node in a precomputed roadmap~\cite{van2008reciprocal}. The local trajectory should be collision-free with other agents and obstacles, at least within the time horizon $\tau$. 

The agent computes the local trajectory by taking into account three factors: its current velocity $\mathbf v_{a_i}$, the observation $\mathbf o_{a_i}$ about the surrounding environment, and its preferred velocity $\mathbf v_{a_i}^{pref}$. In previous work such as~\cite{Berg:ORCA:2011,van2008reciprocal}, the observation $\mathbf o_{a_i}$ consists of the nearby agents' positions, velocities and shapes. However, the estimation of these quantities in the real world requires agent-based recognition and tracking, which is difficult to be implemented reliably. In our method, $\mathbf o_{a_i}$ only includes the raw sensor measurements about the surrounding environment, and thus is more robust and feasible in practice.  

In particular, the agent feeds $\mathbf p_{a_i}$, $\mathbf v_{a_i}$, $\mathbf o_{a_i}$ and $\mathbf v_{a_i}^{pref}$ into a reactive controller $F$, whose output will be parsed as a collision avoidance velocity in the next step:
\begin{equation}
  \mathbf v_{a_i}^+ = F(\mathbf p_{a_i}, \mathbf v_{a_i}, \mathbf o_{a_i}, \mathbf v_{a_i}^{pref}),
\end{equation}
which is then executed by the agent to update its position as
\begin{equation}
  \mathbf p_{a_i}^+ = \mathbf p_{a_i} + \mathbf v_{a_i}^+ \cdot \tau.
\end{equation}
The agent repeats this cycle until arriving at its goal. During the navigation, agents are not allowed to communicate with each other and must make navigation decisions independently, according to the observations collected by their on-board sensors. We do not assume that the agents have perfect sensing about the positions, velocities and shapes of other agents, while such knowledge is usually necessary for previous approaches~\cite{van2008reciprocal,Berg:ORCA:2011,bareiss2015generalized,snape2011hybrid,godoy2015adaptive}. 

The reactive controller $F$ is computed by first converting the observation $\mathbf o_{a_i}$ and $\mathbf v_{a_i}^{pref}$ into the local coordinate frame of the agent $a_i$, and then training a deep neural network $f$ using these data as network inputs. The network solves a multi-class classification problem and outputs a probability distribution which is used to determine the agent's velocity increment $\Delta \mathbf  v_{a_i}$ for safe collision avoidance and making progress towards the goal. The eventual collision avoidance velocity $\mathbf v_{a_i}^+$ is then computed as 
\begin{equation}
\mathbf v_{a_i}^+ = \mathbf v_{a_i} + \Delta \mathbf v_{a_i}.
\end{equation}
We name this deep neural network as the \emph{collision avoidance network} (CANet).

\section{Learning-based Collision Avoidance}
\label{sec:approach}
We begin this section by reviewing the ORCA algorithm, which, with appropriate tuned parameters, is able to produce locally collision-free motions for multiple agents. Next, we describe the details about how to leverage the ORCA algorithm to generate a large training dataset for learning a robust collision avoidance policy in terms of a deep neural network. Finally, we elaborate the network architecture and training details about the collision avoidance policy.

\subsection{A Recap of ORCA}
\label{sec:orca}
In a nutshell, ORCA takes two steps to determine a collision-avoidance velocity $\mathbf v_{a_i}^+$ for an agent $a_i$. First, it computes a set of velocities that form the permitted velocity space for the agent, 
i.e., if choosing a velocity within this space, the agent $a_i$ will not collide with other agents in a time horizon $\tau$. The permitted velocity set is denoted as ${ORCA}^{\tau}_{a_i}$ 
Next, among these permitted velocities, the agent
selects the collision avoidance velocity as a velocity that locates inside the permitted velocity space but is closest to its current preferred velocity $\mathbf v_{a_i}^{pref}$, i.e., 
\begin{equation}
  \mathbf v_{a_i}^+ = \argmin_{\mathbf v \in ORCA^{\tau}_{a_i}} \|\mathbf v - \mathbf v_{a_i}^{pref}\|,
\end{equation}
where $\mathbf v_{a_i}^{pref}$ has been introduced in Section~\ref{sec:prob}.
For good performance, the ORCA's parameters must be tuned carefully during the simulation for different scenarios. A list of the related ORCA parameters is shown in Table~\ref{tab:orca}. While varying these parameters, ORCA presents different collision avoidance behaviors, e.g. agents will be more "shy" if \textsc{neighborDist} is assigned a larger value. For some highly symmetric scenarios, ORCA agents will get stuck by each other without a carefully chosen \textsc{timeHorizon}. Furthermore, if you change the agent's \textsc{protectRadius}, you will need to select new values for all other parameters.

\begin{table}
 \begin{tabularx}{0.5\textwidth}{l|X}
  Parameter & Meaning  \\
   \hline
   \hline
   \textsc{maxSpeed} & the maximum speed of an agent  \\
   \hline
  \textsc{maxNeighbors} & the maximal number of other agents that 
   an agent takes into account in the navigation \\
   \hline
   \textsc{neighborDist} & the maximal distance to other agents that 
   an agent takes into account in the navigation \\
   \hline
   \textsc{ProtectRadius} & the radius of a virtual protection circle centered at the agent where no obstacles shall enter \\
\hline
   \textsc{radius} & the physical radius of an agent \\
   \hline
   \textsc{timeHorizon} & the minimal time horizon for agents to compute collision-free velocities w.r.t. other agents \\
   \hline 
   \textsc{timeHorizonObs} & the minimal time horizon for agents 
   to compute collision-free velocities w.r.t. obstacles \\  
 \end{tabularx}
\caption{The ORCA's parameters.}
\label{tab:orca}
\vspace*{-0.1in}
\end{table}

\subsection{Dataset}
\label{sec:data}
The training of deep neural networks requires a sizable body of training data.
However, directly collecting multi-robot navigation data in the real world could be both challenging and expensive. Thus, in this work we generate training data using the RVO2 simulator\footnote{http://gamma.cs.unc.edu/RVO2/} running the ORCA algorithm with many different configurations in terms of ORCA parameter settings. In this way, the learned policy will behave superior to a simulator with a fixed parameter in term of collision avoidance robustness and efficiency. 

\subsubsection{\textbf{Data Generation}}
\label{sec:data:data-gen}
\begin{figure}[t] 
\centering
\includegraphics[width=0.6\linewidth]{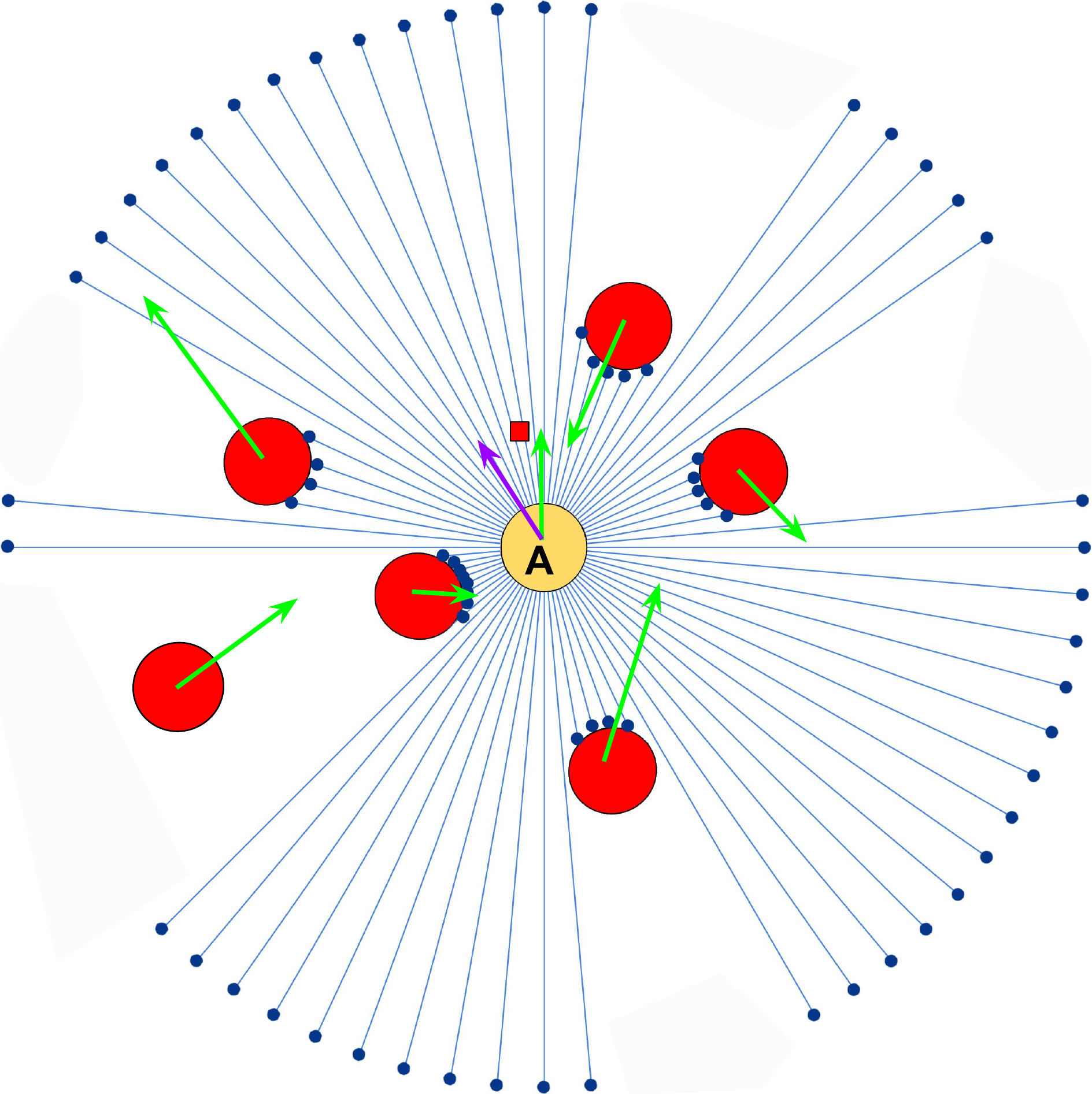}
\caption{Our setup for collecting collision avoidance behaviors of an agent $A$ using the ORCA simulator. The agent $A$ is marked as the yellow circle and locates at the origin. Its current preferred velocity points toward the tiny red square. Around $A$ are several (red) agents with their positions set randomly inside $A$'s neighborhood. We also randomly sample the current velocities of all agents, shown as the green vectors. The blue line segments are the simulated sensor ray cast from $A$, and the blue points are the scan results of the simulated sensor. The purple vector is $A$'s collision avoidance velocity computed by the ORCA algorithm.
}
\label{fig:datacollection}
\vspace*{-0.2in}
\end{figure}

Our setup for the data collection of collision avoidance behaviors is shown in Figure~\ref{fig:datacollection}. Given an agent $A$ whose data is to be recorded, we put it at the origin in the global coordinate space because the agent's absolute position is not important for the collision avoidance. We then sample its preferred velocity $\mathbf v_A^{pref}$ along a random direction. Next, we generate a few agents randomly placed within the agent $A$'s neighborhood of radius $\textsc{neighborDist}$. The velocities of all agents are randomly initialized: the velocity magnitude is sampled from a uniform distribution over the interval $[0, \textsc{maxSpeed}]$, and the direction is also uniformly sampled from $[-\pi, \pi)$. Note that in this setup we do not add any static obstacles.

We repeat the above setup many times with different simulating configurations and generate a large amount of random scenarios. For each scenario, instead of running the simulator many times to generate a sequence, we only execute the simulator one step to generate one frame of collision avoidance data. The reason is the sequence data will have strong correlations with each other, while for training a deep neural network, data items independent with each other are more desirable.

\begin{figure}
\centering
\includegraphics[width=0.7\linewidth]{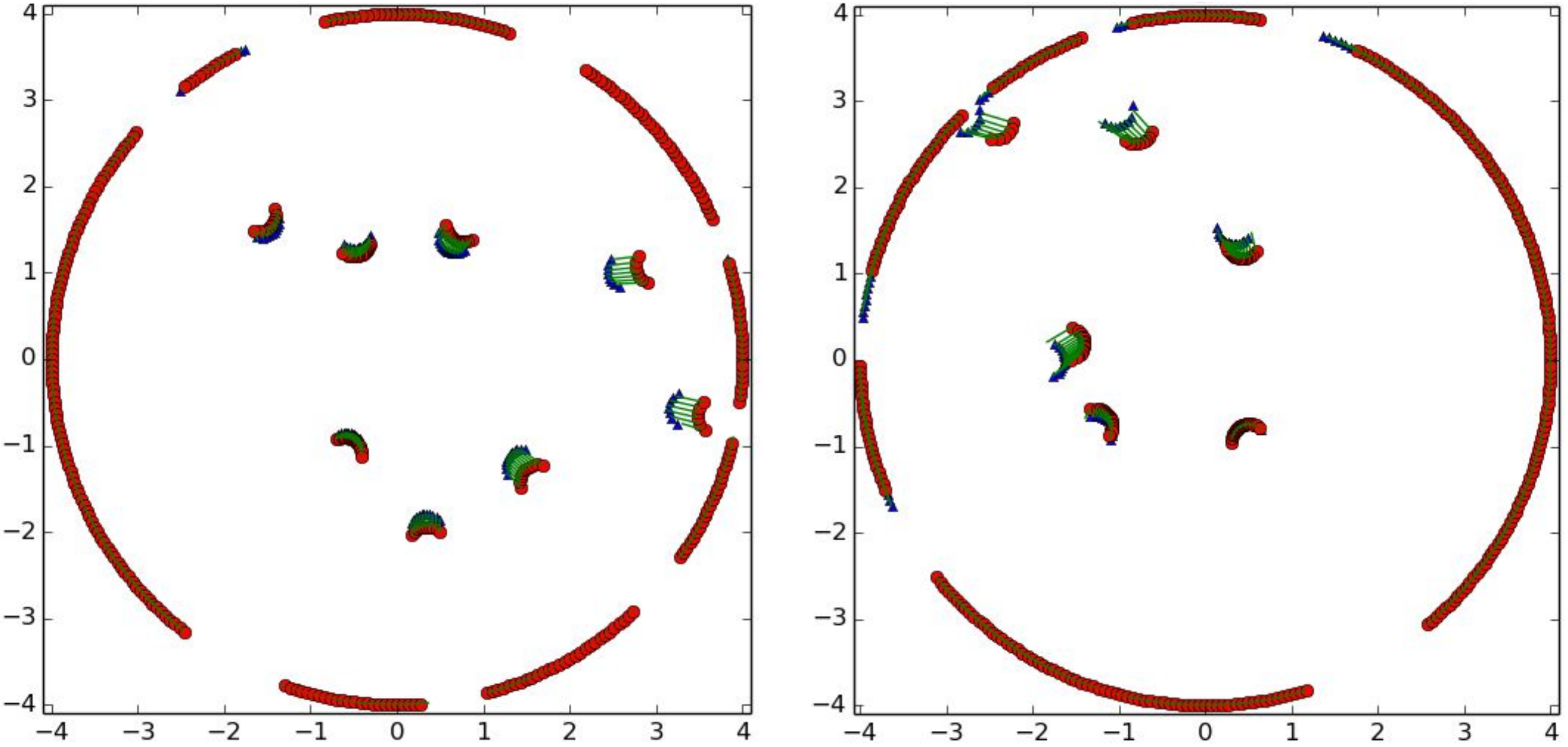}
\caption{The results of non-rigid point matching between scans, which are used to represent the velocities of other agents within an agent's sensing range. The red points are the scan in the previous time step, while the blue points are the scan in the current time step. The green lines illustrate the mapping between two scans. The circle indicate the sensing range of the agent.}
\label{fig:matching}
\vspace*{-0.2in}
\end{figure}

For each frame, we record the agent $A$'s observation $\mathbf o_{A}$ about the surrounding agents, its preferred velocity $\mathbf v_A^{pref}$, and the collision avoidance velocity $\mathbf v_A^+$ calculated by ORCA. To acquire $\mathbf o_{A}$,
we mount a simulated $360$ degree 2D laser scanner at the center of the agent $A$. The simulated scanner has an angular resolution of $1$ degree and a maximum range of $4$ meters. In this way, each scan $\mathbf z_A$ provides $360$ distance values (though in Figure~\ref{fig:datacollection} we only show $72$ scan lines for legibility) ranging from agent radius to the scanner's maximum range. These distance values imply the shapes and positions of other agents in the agent $A$'s surrounding environment. We further infer these neighboring agents' velocities by performing a non-rigid point cloud matching between the current scan and the scan in previous time step using the \emph{coherent point drift} algorithm~\cite{cpd} implemented with the fast Gauss Transform. The matching result estimates the velocity of each point in the current scan, which is denoted as $\dot{\mathbf z}_A$. Two examples of the matched point clouds are shown in Figure~\ref{fig:matching}. 

After collecting $\mathbf o_A = [\mathbf z_A, \dot{\mathbf z}_A]$, $\mathbf v_A^{pref}$ and $\mathbf v_A^+$, we further convert them from the global coordinate space to the local coordinate space fixed at the center of the agent $A$, because the collision avoidance behavior should only rely on the agent's local information. We denote the local observation as $\hat{\mathbf o}_A$, the local preferred velocity as  $\hat{\mathbf v}_A^{pref} = \mathbf v_A^{pref} - \mathbf v_A$, and $\hat{\mathbf v}_A^+ = \mathbf v_A^+ - \mathbf v_A$. In this way, we have prepared the input for the neural network as $[\hat{\mathbf o}_A, \hat{\mathbf v}_A^{pref}]$. The input $\hat{\mathbf o}_A$ has $1080$ dimensions consisting of the scan with $360$ dimensions and its estimated velocity with $720$ dimensions.
The output for the neural network is the label for the velocity $\hat{\mathbf v}_A^+$ in a velocity cluster, as will be discussed later in Section~\ref{sec:data:clustering}.

As described in Table~\ref{tab:orca}, there are seven parameters to be tuned in ORCA, we fix $\textsc{radius}= 0.2$m, $\textsc{maxSpeed} = 3.5$m/s, $\textsc{maxNeighbors} = 10$, $\textsc{neighborDist} = 3.0$m and $\textsc{timeHorizonObs} = 1.0$s for all agents; we vary the $\textsc{protectRadius}$ from set of $\{0.2, 0.5\}$m, and vary $\textsc{timeHorizon}$ from $\{0.5,1.0,2.0\}$s during the data collection. 
We do not vary $\textsc{timeHorizonObs}$ because there is no static obstacle in the training data.
Along with the 
two varied parameters above, another two variables can be changed. The first is the number of $A$'s neighbors, which can vary from $3$ to $10$. The other is the sensor measurement noise, which is a Gaussian noise with a standard deviation ranging between $0.01$ and $0.05$. We generate in total about $310,000$ examples, where each example is a pair in form of $([\hat{\mathbf o}_A, \hat{\mathbf v}_A^{pref}], \hat{\mathbf v}_A^+)$.

\subsubsection{\textbf{Data Cleansing, Augmentation and Preprocessing}} 
\label{sec:data:preprocess}
Before feeding the saved examples to the CANet, we need to first perform some data cleansing, augmentation and preprocessing techniques. For data cleansing, we remove the cases that the agent $A$ updates its position with $\mathbf v_A^{+}$ computed by ORCA but still collides with its neighbors. We then delete the unreasonable outliers by checking whether the speed of collision avoidance velocity $\mathbf v_A^{+}$ is close to $\textsc{maxSpeed}$. For data augmentation, we generate more examples by first adding measurement noises to the inputs and secondly leveraging  symmetries in the collision avoidance scenario along the axis of $\mathbf v_A$, i.e., if we mirror the positions and velocities of the neighboring agents and the preferred velocity along the axis of $\mathbf v_A$, the mirrored version of $\mathbf v_A^+$ will be a valid collision avoidance velocity. As shown in~\cite{krizhevsky2012imagenet,amodei2015deep}, data augmentation technologies can moderate over-fitting and improve the generalization capability of the learned policy. 
It is important to note that data augmentation only adds some redundancy and does not rely on any external source of new information. 

Finally, the training data will be performed standardization (or Z-Score normalization) before being fed to the network.

\subsubsection{\textbf{Collision Avoidance Velocity Clustering}}
\label{sec:data:clustering}
As stated in Section~\ref{sec:prob}, we formulate the computation of the reactive collision avoidance strategy as a multi-class classification problem. In other words, we divide the space of all possible collision avoidance velocities into several classes; and in the runtime the reactive controller will determine which class the collision avoidance velocity should be chosen from, given the sensor observation and the preferred velocity.

\begin{figure} 
\hspace{0.2in}
\begin{subfigure}{0.20\textwidth}
\includegraphics[width=1.0\linewidth]{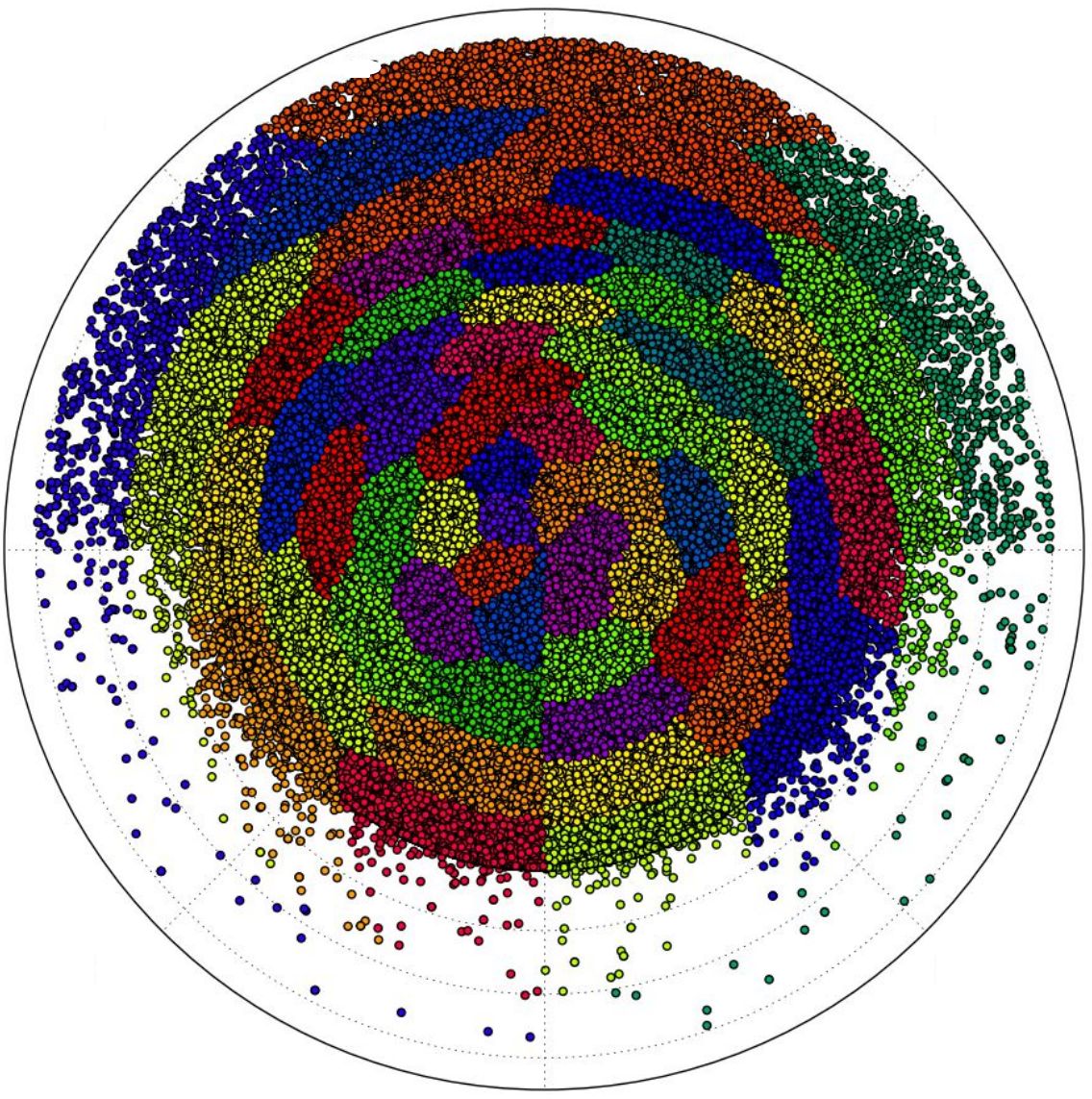}
\caption{$k$-means}
\label{fig:km}
\end{subfigure}
\begin{subfigure}{0.20\textwidth}
\includegraphics[width=1.0\linewidth]{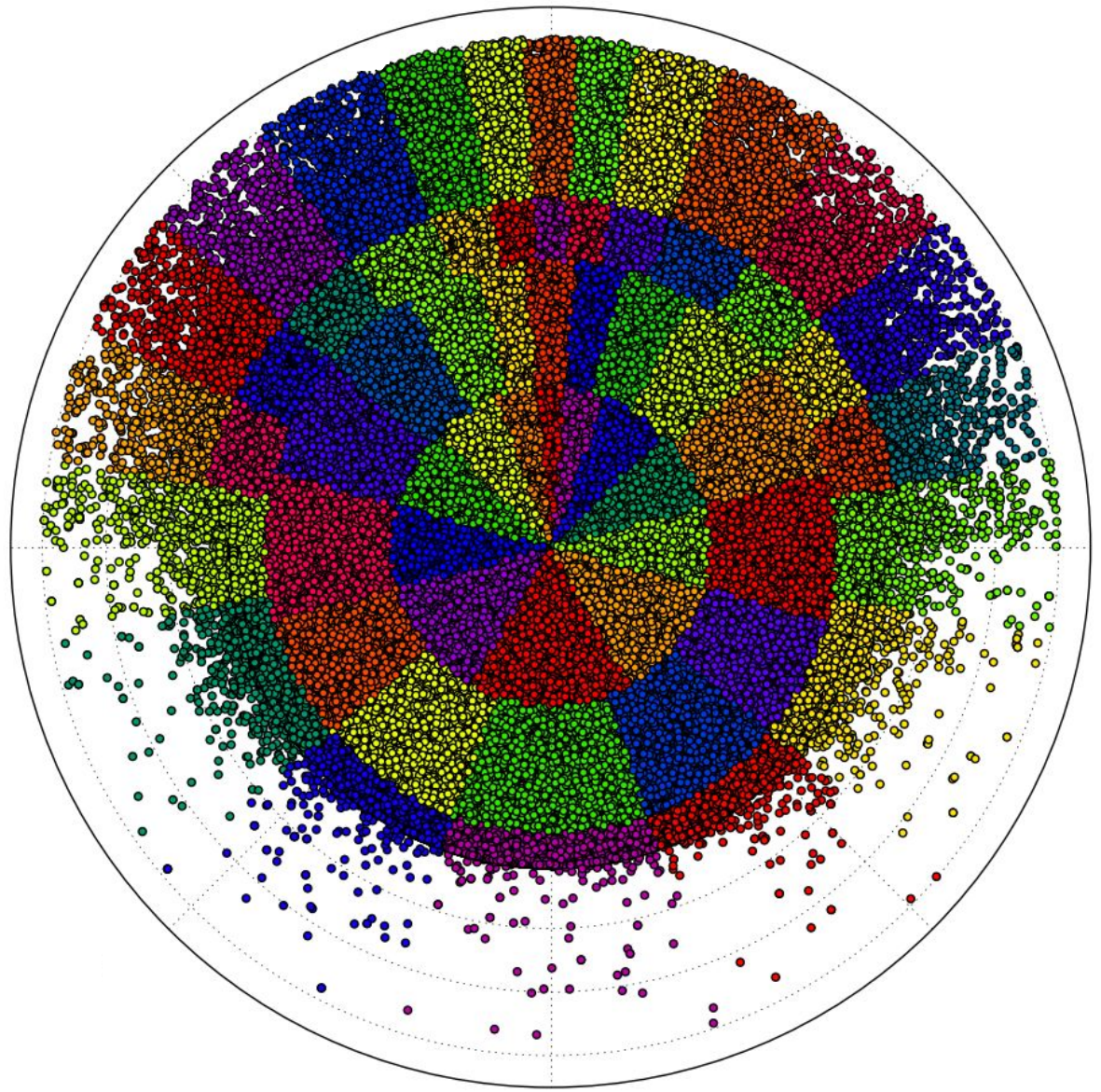}
\caption{manual partition}
\label{fig:manually}
\end{subfigure}
\caption{Partition of the space of collision avoidance velocity. (a) The $k$-means clustering results on the collision avoidance velocities in the collected dataset; (b) A Manually designed partition with 61 classes. Each point in both figures represents a collision avoidance velocity $\mathbf v^+$, and its class label is determined by the point color.}
\label{fig:clustering}
\vspace*{-0.1in}
\end{figure}

To generate a reasonable partition for the space of collision avoidance velocity, we first perform a $k$-means clustering on the $\mathbf v_A^+$ and use the clustering result shown in Figure~\ref{fig:km} as a reference, we manually design a partition with 61 classes as shown in Figure~\ref{fig:manually}.

We choose to not model the computation of collision avoidance velocity as a regression problem because the $l_2$ loss function for regression tasks usually is more fragile to outliers~\cite{Belagiannis:ICCV:ROD}. In addition, it is more desirable to output a probability about selecting a collision avoidance velocity given a noisy sensor measurement, but the regression only generates a single velocity output with no indication about the confidence. 

\subsection{Collision Avoidance Network} 
\label{sec:net}
\begin{figure}
\centering 
\includegraphics[width=0.8\linewidth]{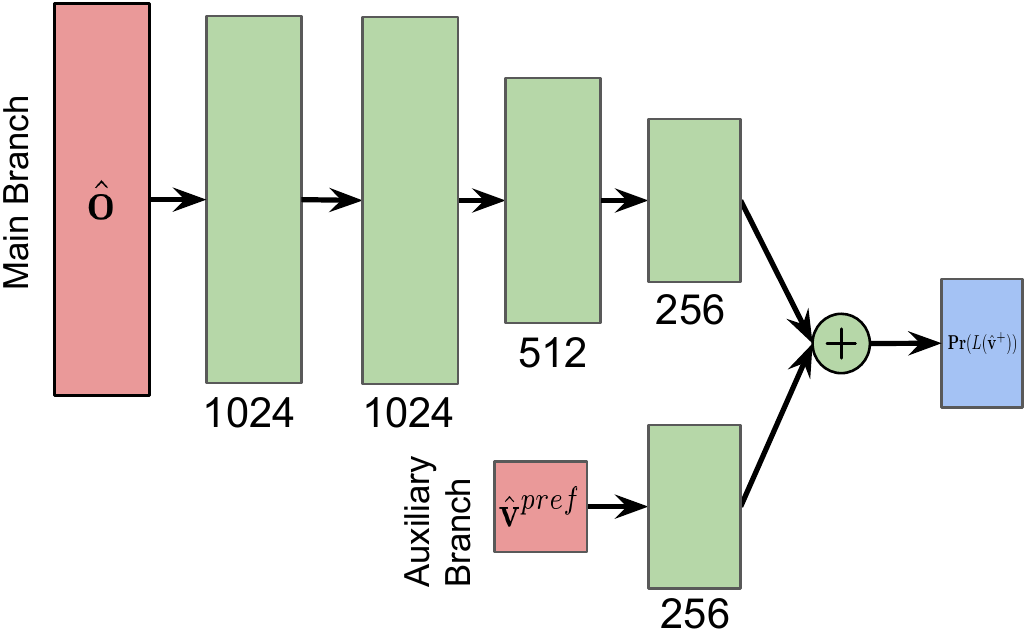} 
\caption{The architecture of the collision avoidance network CANet. The network has the local observation $\hat{\mathbf o}$ and the local preferred velocity $\hat{\mathbf v}^{pref}$ as inputs, and outputs a probability distribution $\mathbf{Pr}(L(\hat{\mathbf v}^+))$ that can be parsed to the collision avoidance velocity, where $L(\hat{\mathbf v}^+)$ is the class label for each $\hat{\mathbf v}^+$.} 
\label{fig:net} 
\vspace*{-0.2in}
\end{figure}

The CANet is a two-branch multilayer perceptron (MLP) and its architecture is as summarized in Figure~\ref{fig:net}. Following is the details about the different components of this network:  
\subsubsection{\textbf{Architecture}}
The CANet has two branches. The input of the main branch is the agent's observation $\hat{\mathbf o}_A$ and the input of the auxiliary branch is $\hat{\mathbf v}_A^{pref}$, which are both described in Section~\ref{sec:data:data-gen}. The output of the CANet is a probability distribution over the velocity classes which will be parsed to the collision avoidance velocity $\hat{\mathbf v}^+$ for updating the agent's position. In the main branch, there are four fully connected hidden layers after the input layer, and these layers consist of $1024$, $1024$, $512$ and $256$ rectified linear units (ReLUs) respectively. A dropout layer with probability $0.2$ is applied after each of these hidden layers. The auxiliary branch has only one fully connected hidden layer with $256$ ReLUs. The two branches are merged by concatenating the fourth layer of the main branch with the hidden layer of the auxiliary branch. This merged layer is then followed by a fully connected layer with one neuron per class, activated by a softmax function. We use the cross-entropy loss function during the training stage.  

\subsubsection{\textbf{Training}}
We randomly split the dataset into ten stratified folds preserving the percentage of samples for each class and report results using 10-fold cross-validation. 
We train our network for about 5 hours until convergence on a single Nvidia Titan X GPU, using stochastic gradient descent (SGD) with the momentum of $0.9$. We use a base learning rate of $0.1$ and a decay rate of $0.0002$. The network is trained for the maximal $300$ epochs with early stopping using a batch size of $64$. The classification accuracy on the test dataset is $33.435\%$ ($\pm 0.354\%$) and $64.106\%$ ($\pm 0.313\%$) on the training set. 

\subsubsection{\textbf{Collision Avoidance Velocity}} CANet will output a distribution over the collision avoidance velocity $\mathbf{Pr}(l=L(\hat{\mathbf v}^+))$, where $L(\hat{\mathbf v}^+)$ is the class label for each $\hat{\mathbf v}^+$. In this work, we determine the actual collision avoidance velocity using a simple method. We first choose the class $l$ with the highest probability and then perform random sampling inside the class around the class centroid. Next, we compute the safety margins (i.e., closest distance to obstacles) when the agent applies these sampled velocities as the collision avoidance velocity in the time horizon $\tau$, and choose the velocity with the maximum safety margin as our result. If this velocity will make the agent collide with obstacles (i.e., the minimum safety margin is negative), we will slow down the velocity accordingly.

\section{Experiments and Results}
\label{sec:exp}

This section presents experiments and results of the proposed framework. We have evaluated this framework in various simulated scenarios and compared it to ORCA. We have also tested our method on a real multi-robot system. 

\subsection{Experiment Setup}
\subsubsection{\textbf{Scenarios}} 
\begin{figure} 
\centering
\includegraphics[width=0.8\linewidth]{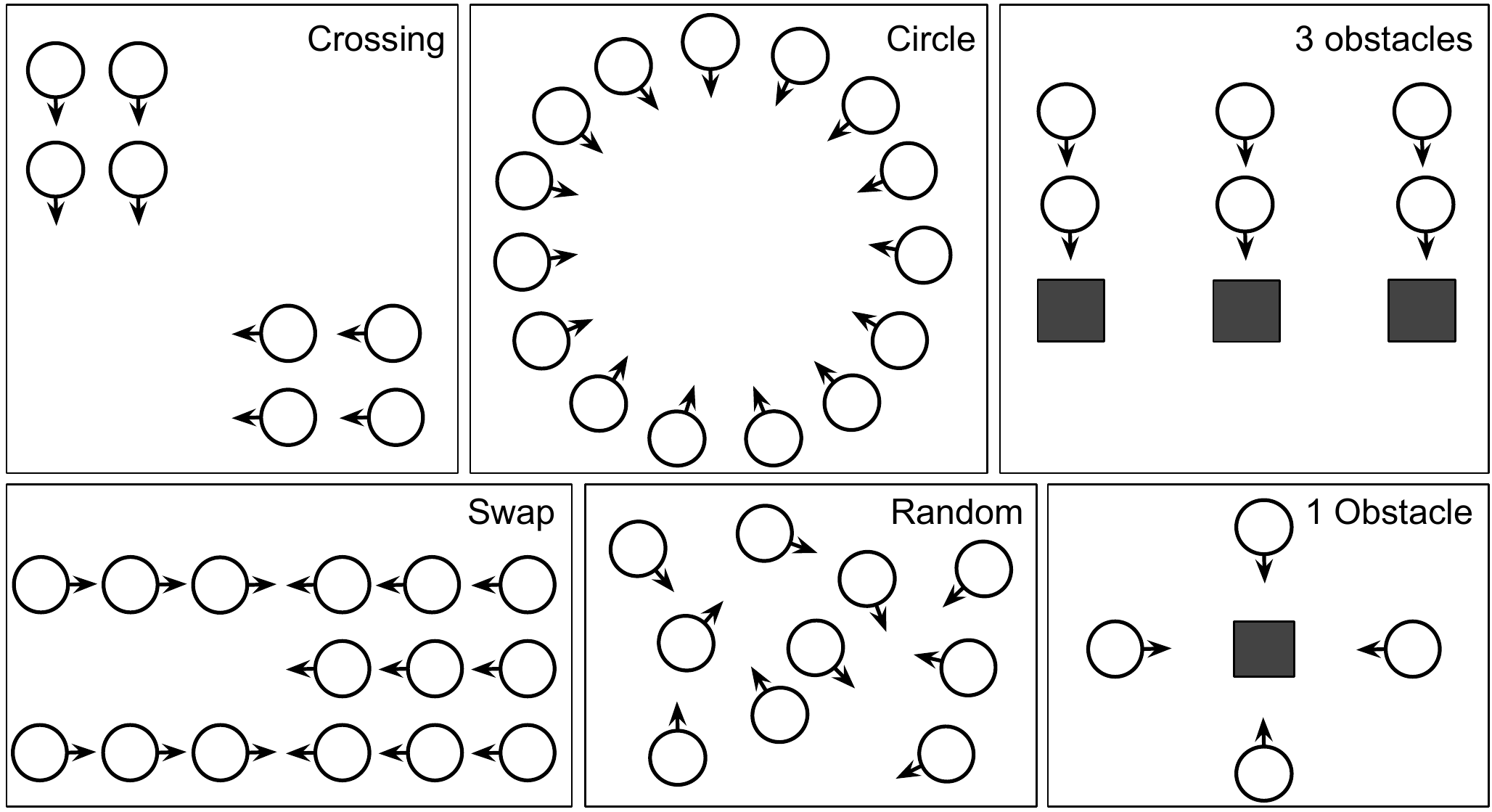}
\caption{Six scenarios used to compare the navigation performance of our learned policy and the ORCA policy. 
}
\label{fig:scenarios}
\vspace*{-0.15in}
\end{figure}

We evaluated our learned policy on six different scenarios with different number of agents (as shown in Figure~\ref{fig:scenarios}). Note that
the test scenarios \textit{3 Obstacles} and \textit{1 Obstacle} have static obstacles, which never appear in any data collection scenarios for training the CANet. In addition, since the trained policy outputs the collision avoidance velocity in a random manner, its performance is averaged over 20 simulations. 

\begin{figure}[!h]
\centering
\includegraphics[width=0.8\linewidth]{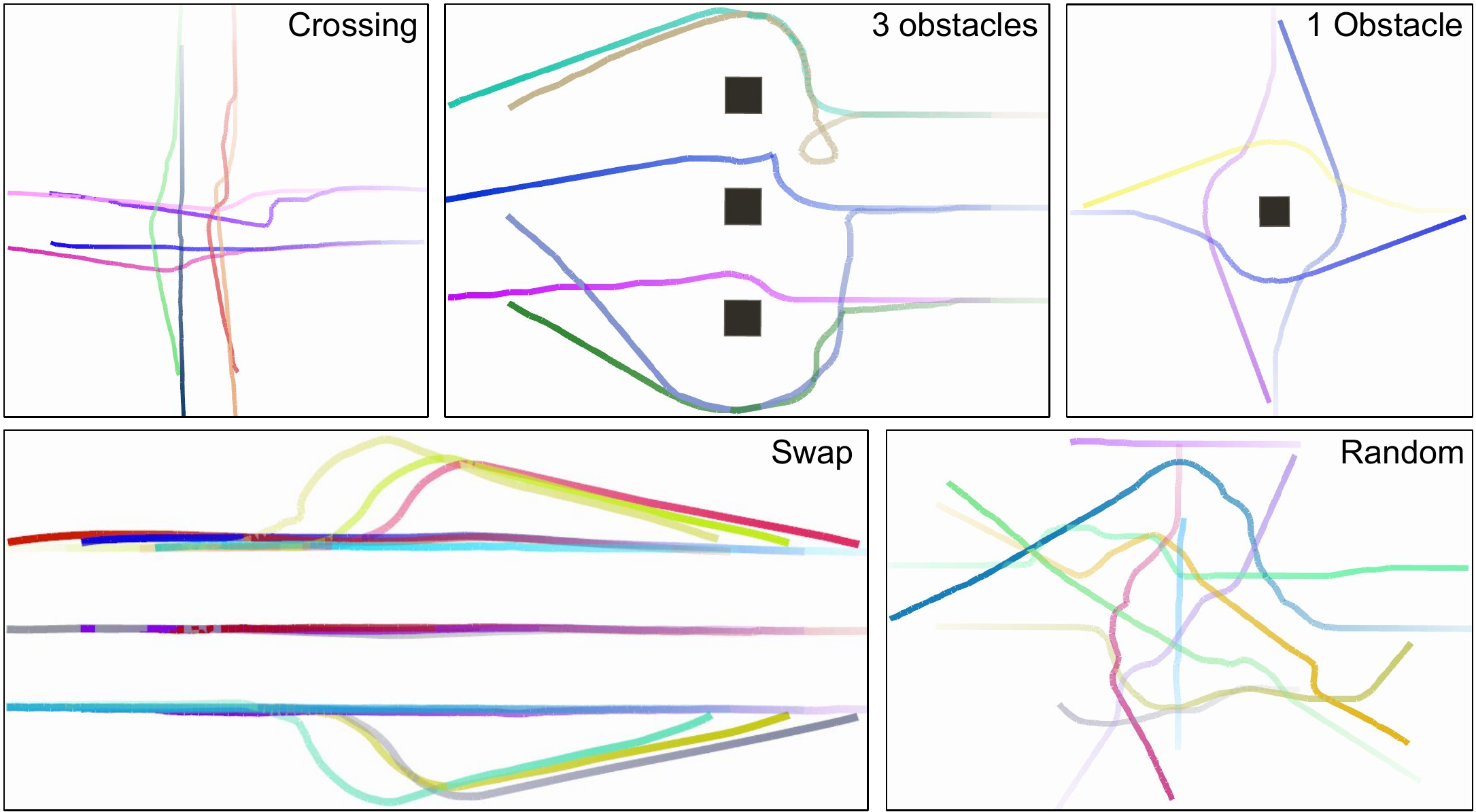}
\caption{Trajectories of five scenarios using the learned policy.} 
\label{fig:paths}
\vspace*{-0.2in}
\end{figure}

We compared the performance of learned policy with the ORCA policy. Most parameters of the ORCA policy are set to be the same as the values used in the data generation for the learned policy (as stated in Section~\ref{sec:data:data-gen}), but we tuned some parameters to optimize ORCA's performance. In particular, to obtain the best performance of ORCA, we 
change $\textsc{timeHorizonObs}$ to $10.0$s for scenarios with static obstacles and tune $\textsc{timeHorizon}$ for different scenarios. 
In each simulation, the performance of ORCA is evaluated with two different $\textsc{protectRadius}$ values  $0.2$m and $0.5$m.
The time step size $\tau$ of the sensing-acting cycle is set to $0.1$s. The detailed description for each test scenario is as follows:
\renewcommand{\labelitemi}{\textbullet}
\begin{itemize}
\item \textit{Crossing}: agents are separated in two groups, and their path will intersect in the bottom left corner;
\item \textit{Circle}: agents are initially located along a circle and each agent's goal is to reach its antipodal position;
\item \textit{Swap}: two groups of agents moving in opposite directions swap their positions;
\item \textit{Random}: agents are randomly initialized in a cluttered environment and are assigned random goals; 
\item \textit{3 Obstacles}: six agents move across three obstacles;
\item \textit{1 Obstacle}: four agents initialized on a circle move towards their antipodal positions, and an obstacle is located at the center. 
\end{itemize}
The trajectories generated using the learned navigation policy are shown in Figure~\ref{fig:circle} and Figure~\ref{fig:paths}.

\subsubsection{\textbf{Performance Metrics}} To compare the performance of our framework and ORCA quantitatively, we use the following performance metrics: 
\renewcommand{\labelitemi}{\textbullet}
\begin{itemize}
\item \textit{Total travel time}: the time taken by the last agent to reach its goal;
\item \textit{Total distance traveled}: the total distance traveled by all agents to reach their goals.
\item \textit{Safety margin}: the agent's closest distance to other agents and static obstacles;
\item \textit{Completion}: if all agents reach their goals within a time limit without any collisions, the scenario is successfully completed. 
\end{itemize}

\subsection{Quantitative Comparisons}
\begin{figure} 
\centering
\includegraphics[width=1\linewidth]{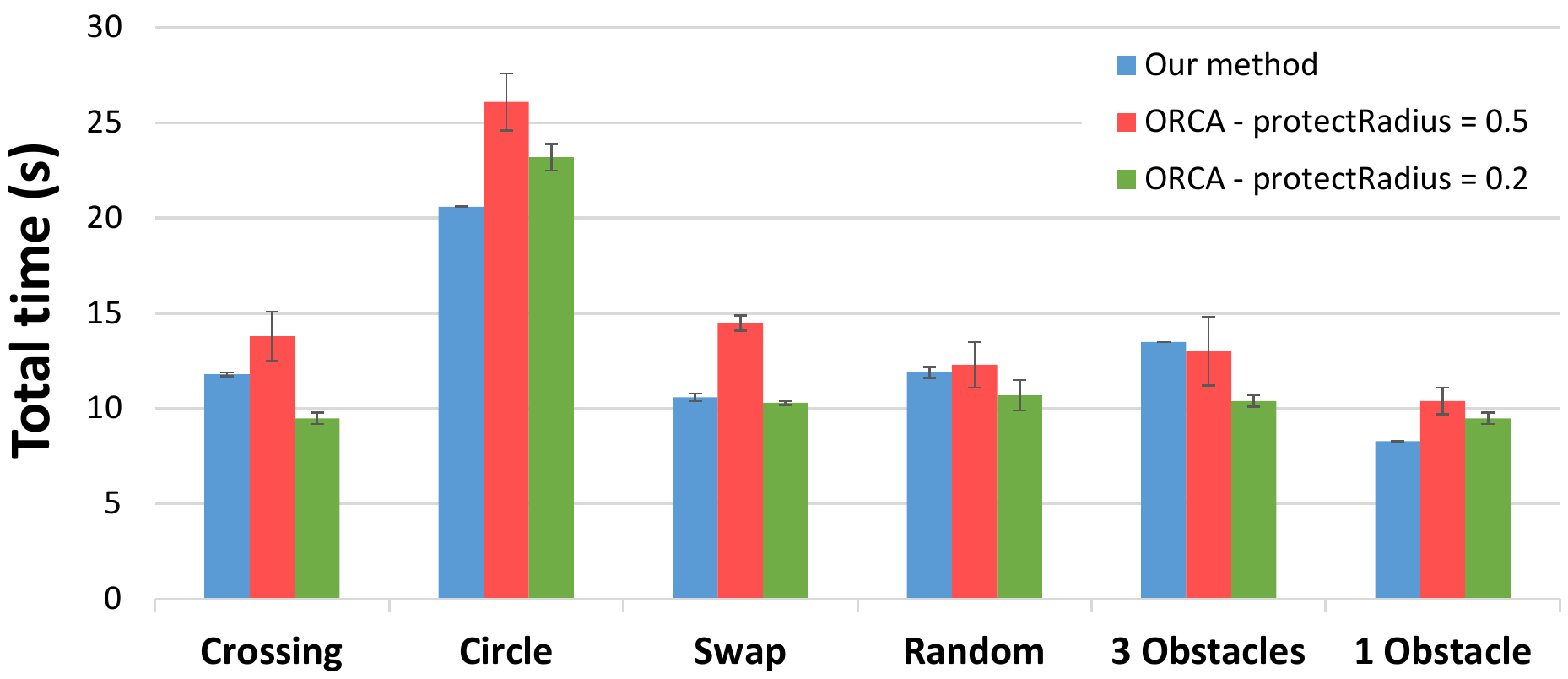}
\caption{Total time of our method and ORCA in all scenarios.}
\label{fig:time}
\end{figure}

\begin{figure} 
\centering
\includegraphics[width=1\linewidth]{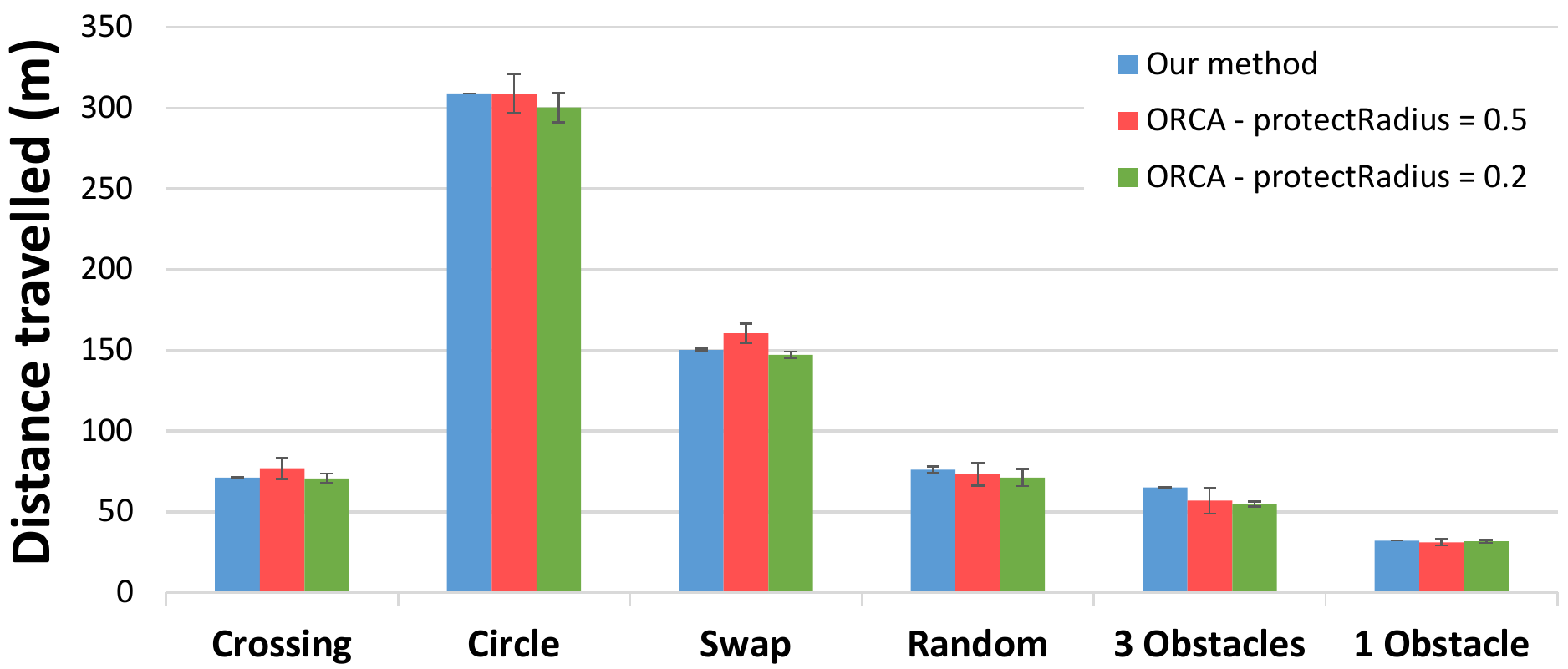}
\caption{Distance traveled of our method and ORCA in all scenarios.}
\label{fig:distance}
\end{figure}

In Figure~\ref{fig:time} and~\ref{fig:distance}, we measure two metrics -- \textit{total travel time} and \textit{total traveled distance} -- to evaluate the performance of our approach and ORCA.
We can observe that when comparing with the ORCA policy with $\textsc{protectRadius}=0.5$, the learned policy provides better or comparable performance in terms of navigation duration and trajectory length. In most scenarios, the ORCA policy with $\textsc{protectRadius}=0.2$ has shorter navigation time and trajectory length than the learned policy. This is because the ORCA policy with $\textsc{protectRadius}=0.2$ is very aggressive and allows a small safe margin during the navigation, as shown in Table~\ref{tab:safety-margin}. Both our learned policy and the ORCA policy with $\textsc{ProtectRadius}=0.5$ try to keep a large enough margin with nearby agents/obstacles. 
The difference is that the ORCA policy with $\textsc{ProtectRadius}=0.5$ uses the protect radius parameter to keep a hard margin: no obstacles/agents should get closer to the agent than $\textsc{ProtectRadius} - \textsc{radius} = 0.5 - 0.2 = 0.3$m, and this constraint may be too conservative in a cluttered scene. 

Instead, our method learns the preference for margin implicitly from the data and is able to keep the clearance in an adaptive manner: in cluttered situations, the agents can endure a small safety-margin while in an open space, the agents will tend to keep a large safety-margin. For instance, the safety margin in the \emph{3 Obstacles} scenario is smaller than in the \emph{1 Obstacle} scene as shown in Table~\ref{tab:safety-margin}, because the former is more cluttered.

\begin{figure}[h] 
\begin{subfigure}{0.2\textwidth}
\includegraphics[width=1.0\linewidth]{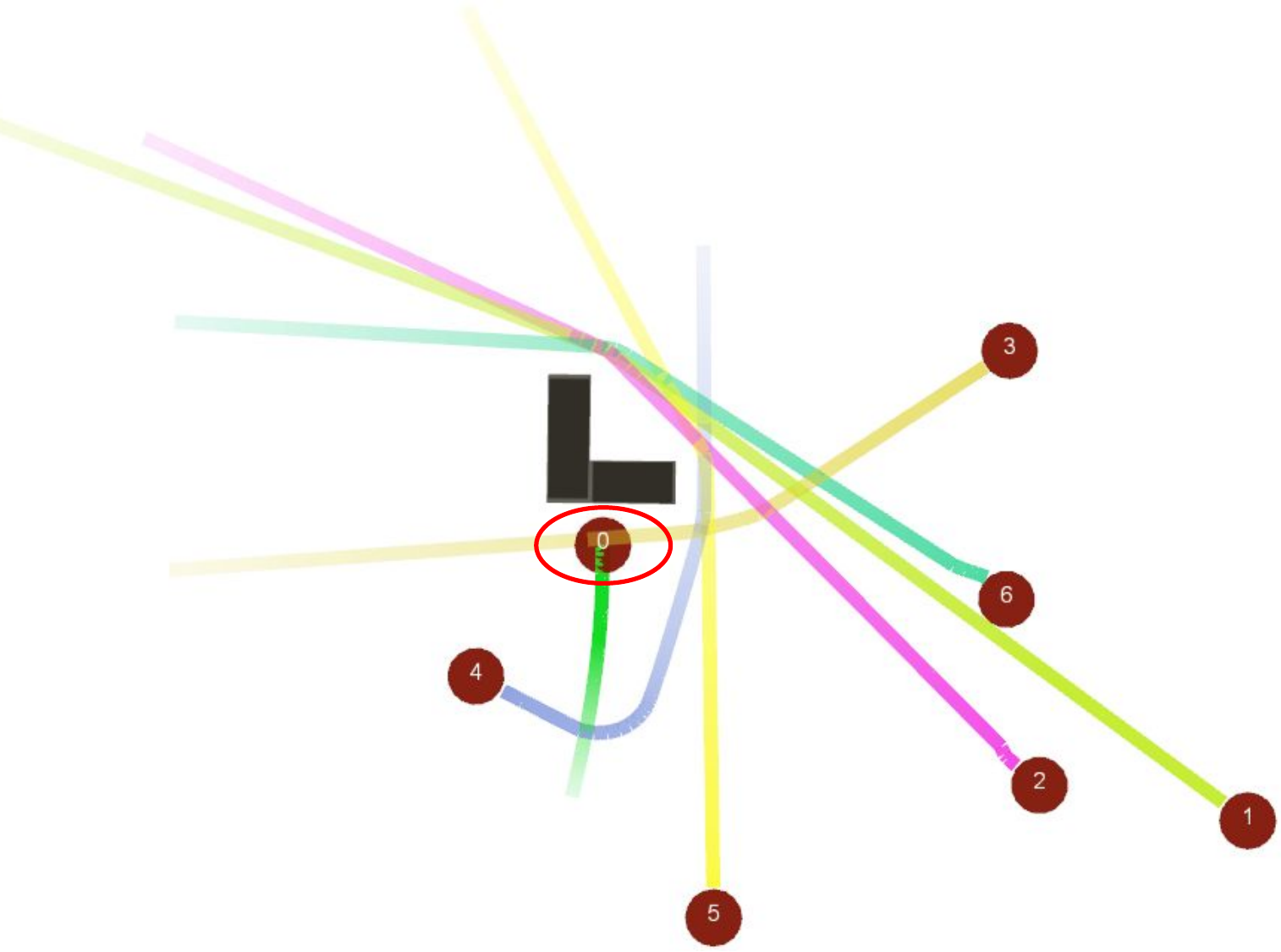}
\caption{ORCA}
\label{fig:stuck-rvo}
\end{subfigure}
\begin{subfigure}{0.2\textwidth}
\includegraphics[width=1.0\linewidth]{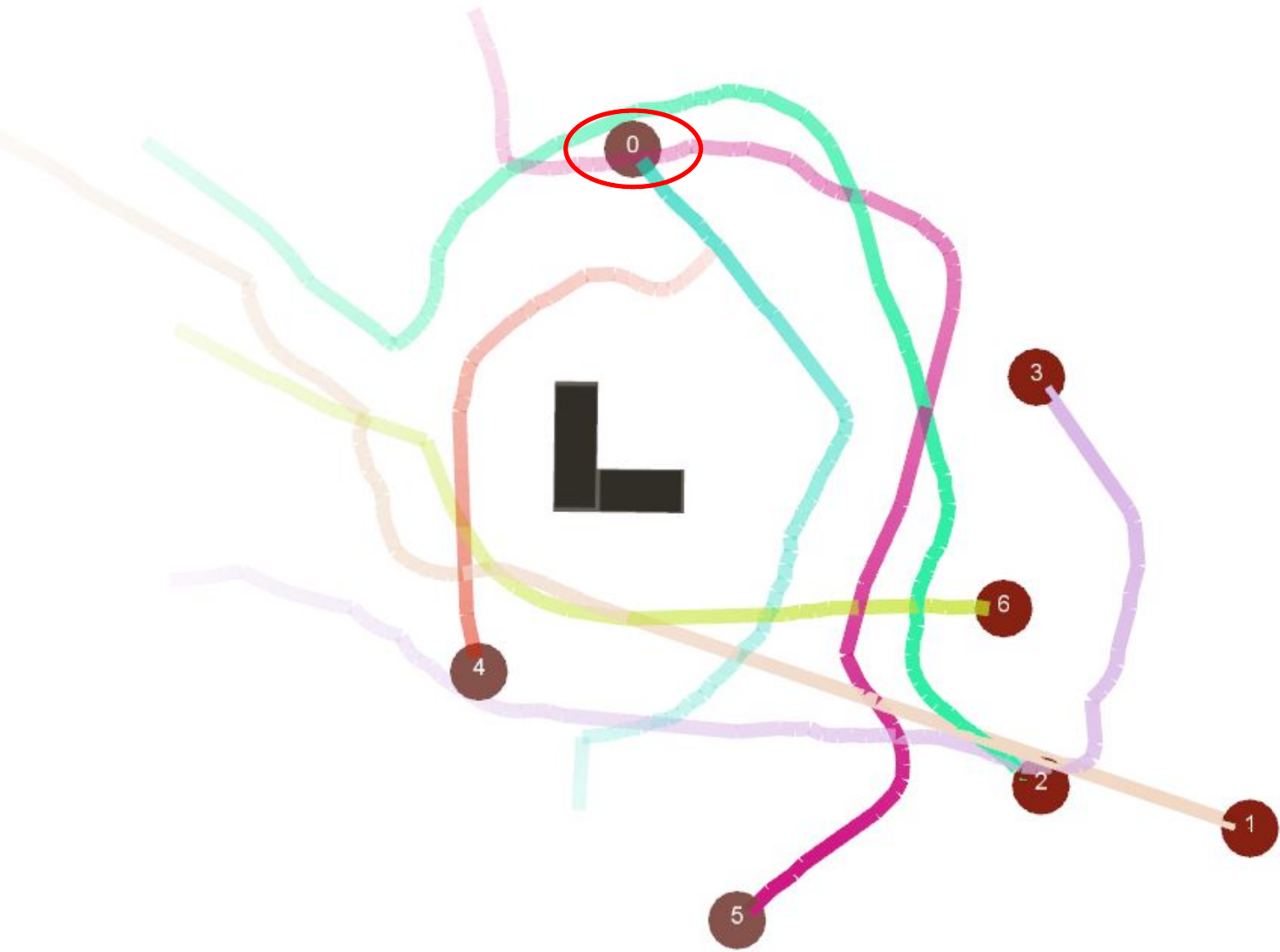}
\caption{Our method}
\label{fig:stuck-dnn}
\end{subfigure}

\caption{For the given scenario with static obstacles,
one agent gets stuck when using the ORCA policy while all agents reach their goals with the learned policy.}
\label{fig:stuck}
\vspace*{-0.2in}
\end{figure}

We also set up a more challenging scenario with an L-shape static obstacle at the center (shown in Figure~\ref{fig:stuck}) and measure the \textit{Completion} metric. We randomly generate $100$ initial states where all agents are randomly placed and they are assigned with appropriate random goals. We then compare our method and ORCA by counting the number of failures, i.e., some agents do not reach their goals or severely collide with other agents/obstacles during the runtime. ORCA has a failure rate of $15\%$ while our learned policy only has $2\%$. Figure~\ref{fig:stuck} shows a \textit{stuck} case for ORCA while our learned policy can complete it successfully.

\begin{table*}
 \begin{tabularx}{1\textwidth}{l|X|X|X|X|X|X}
  	Safety Margin (min/ave) & Crossing & Circle & Swap & Random & 3 Obstacles & 1 Obstacle  \\
   \hline
   Our method &  0.028 / 0.197 & 0.281 / 0.281 & 0.209 / 0.365 & 0.171 / 0.334 & 0.012 / 0.188 &0.108 / 0.154\\
   \hline
   ORCA - \textsc{protectRadius}=0.5 & 0.300 / 0.375 &	0.262 / 0.297 &	0.276 / 0.295 &	0.297 / 0.299 &	0.299 / 0.301 &	0.300 / 0.365
 \\
   \hline
   ORCA - \textsc{protectRadius}=0.2 & 0.000 / 0.003 & -0.016 / -0.004	& 0.000 / 0.098 &	0.000 /  0.136  &	0.000 / 0.000 &	0.004 / 0.004 \\
 \end{tabularx}
\caption{The minimum and average safety margins for agents when using our learned policy and ORCA policy with $\textsc{protectRadius}=0.5$ and $\textsc{protectRadius}=0.2$ in all six scenarios.}
\label{tab:safety-margin}
\vspace*{-0.2in}
\end{table*}

\subsection{Generalization}

An interesting phenomenon while using the learned policy is that in a highly symmetrical scenario like \textit{Circle}, the agents will present certain cooperative behaviors since all agents are using the same learned policy. For instance, a cooperative rotation behavior is shown in Figure~\ref{fig:circle-dnn} where agents starts to rotate at the same pace when they are close to each other. While for ORCA (as shown in Figure~\ref{fig:circle-rvo}), each agent passes the central area by itself without any collective behaviors and some agents yields jerky motions. 

\begin{figure}[h] 
\centering
\begin{subfigure}{0.22\textwidth}
\includegraphics[width=3.5cm,height=3.5cm]{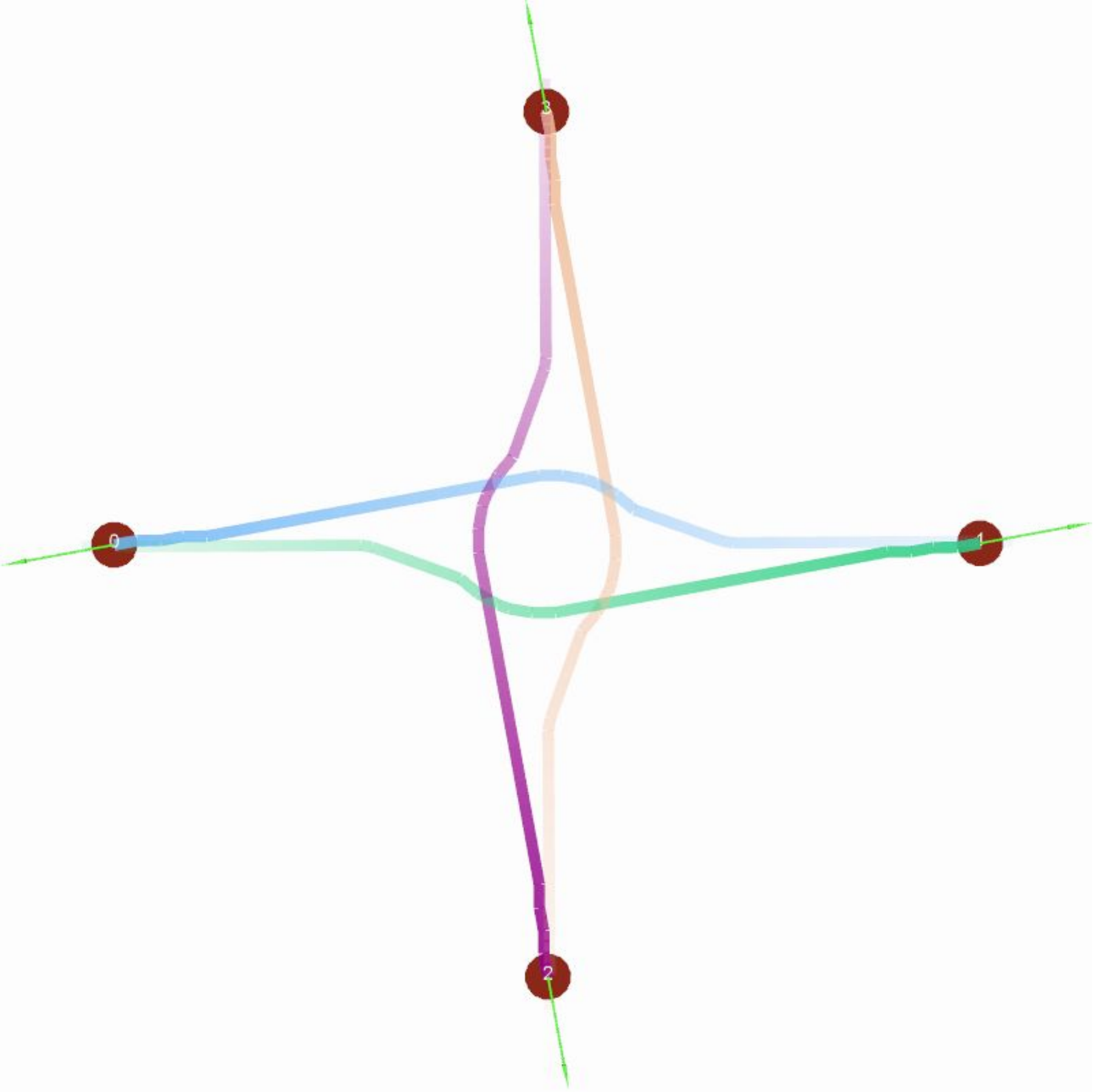}
\caption{Ours -- same $\textsc{radius}$}
\label{fig:ours-same}
\end{subfigure}
\begin{subfigure}{0.22\textwidth}
\includegraphics[width=3.5cm,height=3.5cm]{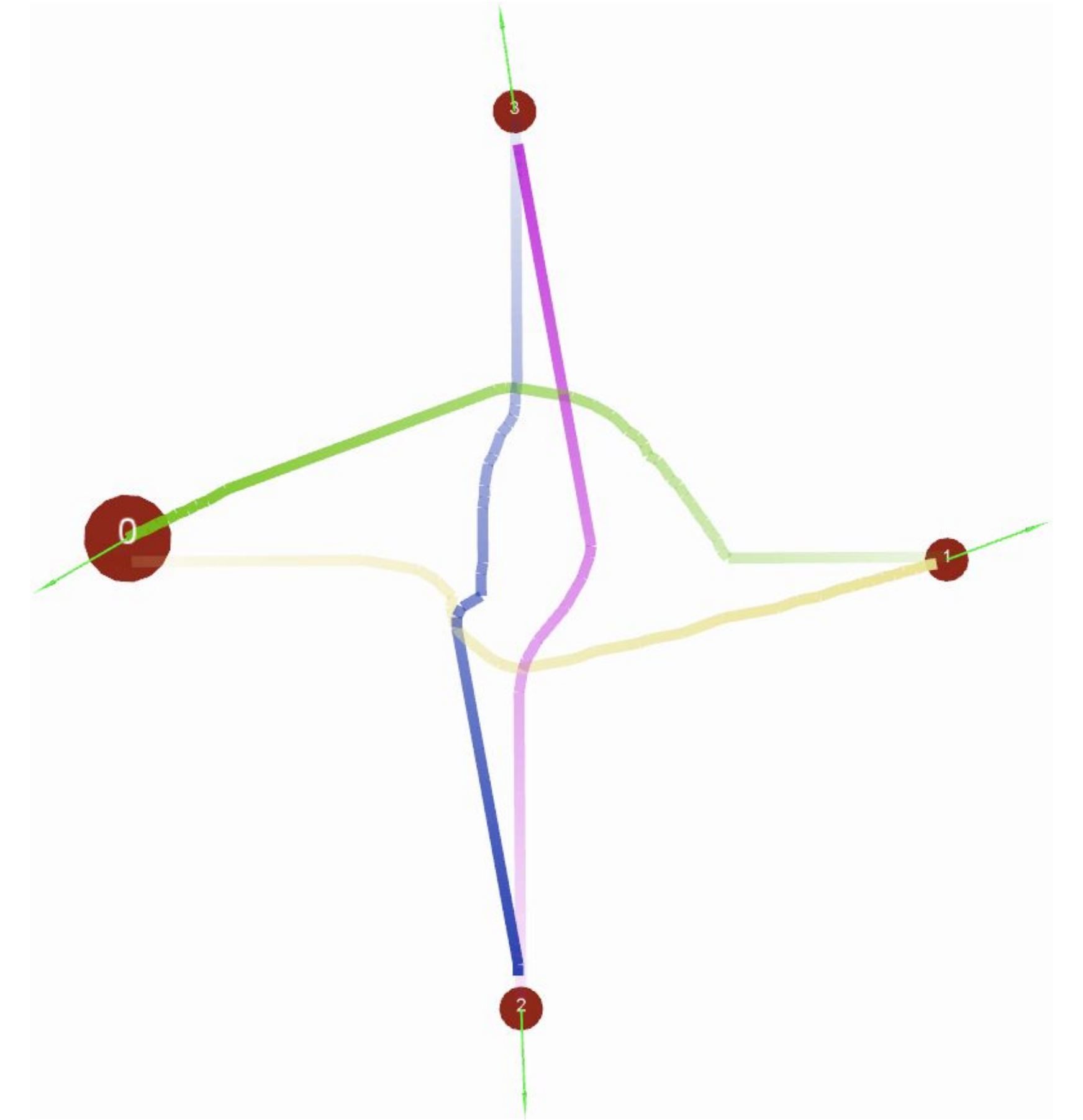}
\caption{Ours -- different $\textsc{radius}$}
\label{fig:ours-diff}
\end{subfigure} \\
\centering
\begin{subfigure}{0.215\textwidth}
\includegraphics[width=3.325cm,height=3cm]{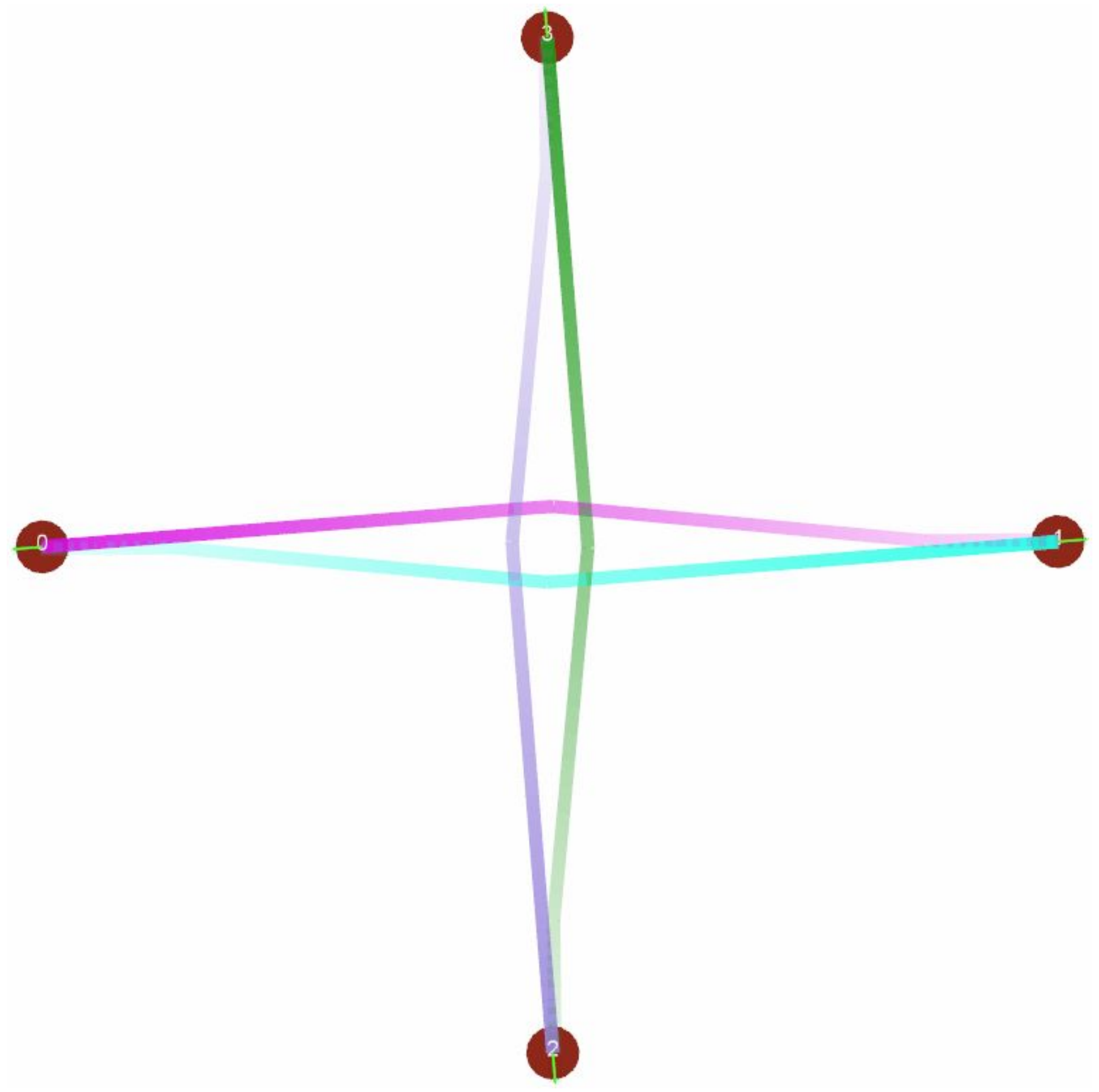}
\caption{ORCA -- same $\textsc{radius}$}
\label{fig:rvo-same}
\end{subfigure}
\begin{subfigure}{0.215\textwidth}
\includegraphics[width=3.325cm,height=3cm]{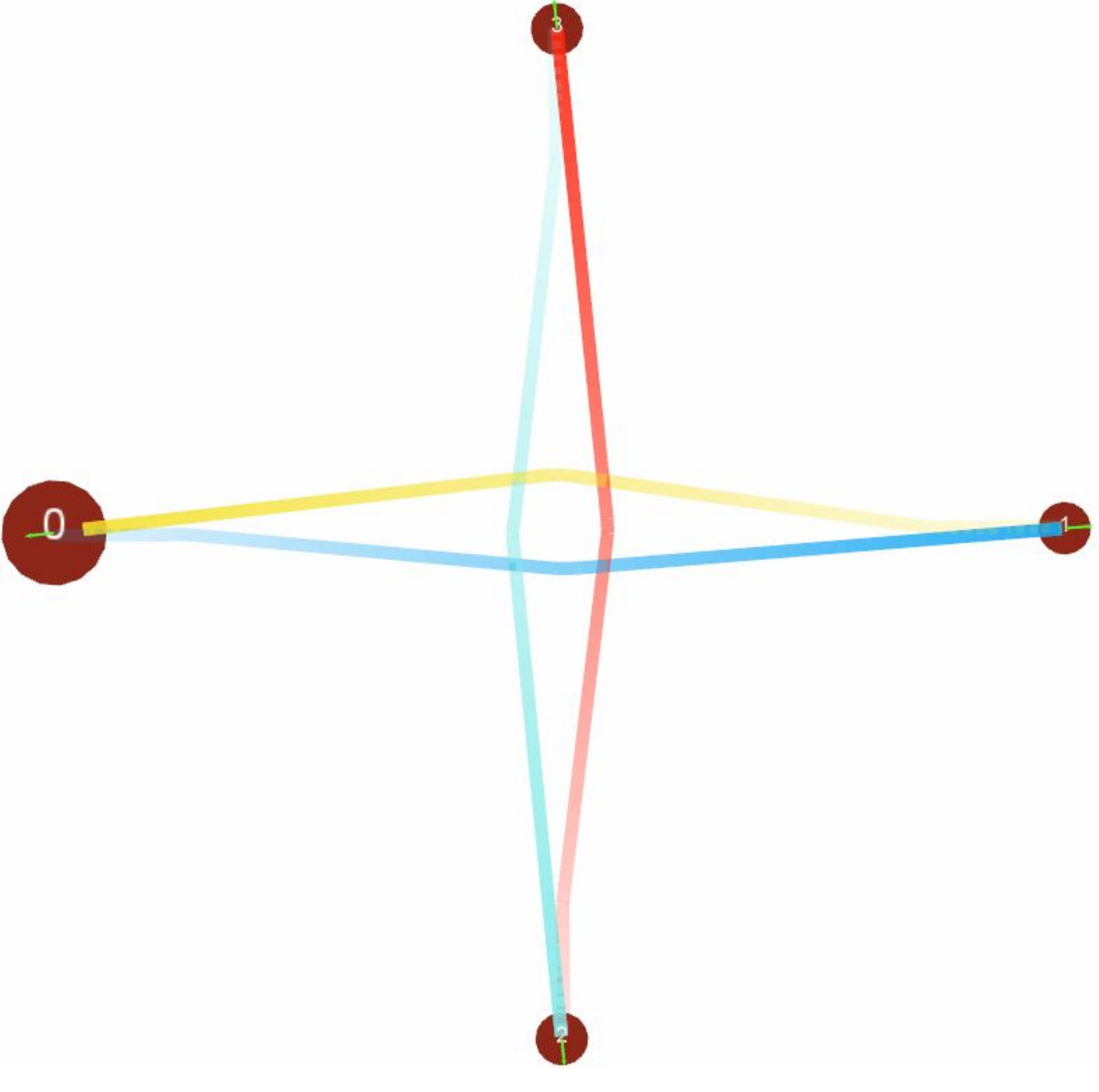}
\caption{ORCA -- different $\textsc{radius}$}
\label{fig:rvo-diff}
\end{subfigure}
\caption{Our method generalizes well to situations where agents have different physical sizes. Four agents have the same size in (a) and have different $\textsc{radius}$ in (b). In (b), one of four agents has a larger size than others. For ORCA (c) and (d), the paths of different-sized agents and same-sized agents are similar, since ORCA explicitly knows all agents' radii and velocities before calculating a steering command.}
\label{fig:size}
\vspace*{-0.1in}
\end{figure}

The good generalization capability is another notable feature of our method. The learned policy's performance  in scenarios with static obstacles demonstrates that it generalizes well to handle to previously unseen situations. In addition, we also evaluate the learned policy in the \textit{Circle} scenario with four different-sized agents. In Figure~\ref{fig:ours-same}, agents with the same size have identical paths as they take the same strategy to avoid collision with each other. When one agent gets bigger (as shown in Figure~\ref{fig:ours-diff}), it will deviate more from the original path to generate safe movements and this causes other agents on its path to adjust navigation behaviors accordingly. Please note that this experiment (Figure~\ref{fig:rvo-same} and~\ref{fig:rvo-diff}) does not reveal the generalization of ORCA since ORCA always knows all agents' radii before computing the collision-free velocity.

\begin{figure} 
\centering
\includegraphics[width=1\linewidth,height=4.5cm]{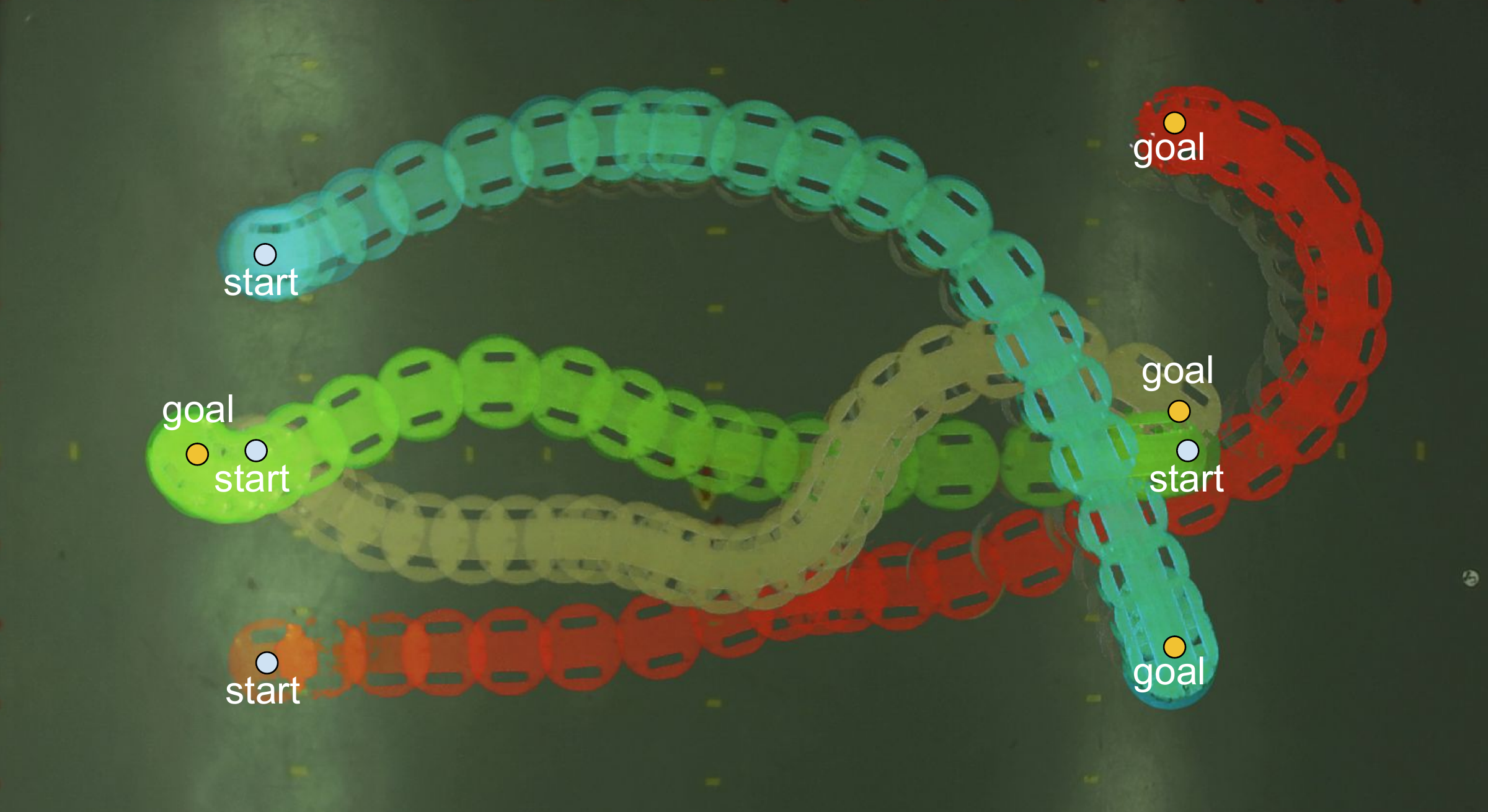}
\caption{A real-robot experiment in 1 vs. 3 scenario. We use different colors to distinguish the trajectories of different agents.}
\label{fig:real-robot}
\vspace*{-0.2in}
\end{figure}

We have also demonstrated the proposed method on real robots where each robot is mounted with a Hokuyo URG-04LX-UG01 2D lidar sensor. In Figure~\ref{fig:real-robot}, four robots, three on the right side and one on the left side, are moving to their antipodal positions. As we observe, each robot can effectively avoid collisions with other robots during the navigation in a complete distributed manner. In addition, our system does not require any AR tags and/or additional motion capture systems to offer each agent with the position and/or velocity information about the other agents. In this way, our system can achieve real decentralized multi-agent navigation without any centralized components. 

\section{Conclusion and Limitations}
\label{sec:conclusion}

This paper is our first step toward learning a reactive collision avoidance policy for efficient and safe multi-agent navigation. By carefully designing the data collection process and leveraging an end-to-end learning framework, our method can learn a deep neural network based collision avoidance policy which demonstrates an advantage over the state-of-the-art ORCA policy in terms of ease of use (no parameter tuning), success rate, and navigation performance. 
In addition, even though being trained over dataset with only identical moving agents, our learned policy generalizes well to various unseen situations, including agents with different sizes and scenarios with static obstacles. 

The proposed method has some limitations. First, at the current stage, we are training a vanilla multilayer perceptron as the collision avoidance policy. As can be observed from the classification accuracy, the model does not completely fit the training data (the accuracy on training set is around $64\%$), thus there is still great potential for getting the model improved. Second, we did not add any static obstacles during training data generation, and therefore our model may not perform well in some challenging scenarios with obstacles (e.g., agents pass through a narrow hallway, multiple agents exit a room through a narrow doorway). These challenging tasks can be solved by combining our method with the cutting-edge deep reinforcement learning techniques, which will further improve agents’ navigation performance.

Besides the combination with reinforcement learning, there are many other exciting avenues for the future work, such as the extension to vehicles with complex dynamics (e.g., the quadrotors), how to directly leverage 2D/3D camera sensors, and most importantly, to make the entire framework work reliably in real systems (e.g., the automated warehouse) with a large number of agents.

{\small
\bibliographystyle{IEEEtran}
\bibliography{references}
}

\end{document}